\documentclass[twoside,leqno,twocolumn]{article}  

\newif\ifapx
\apxtrue 

\usepackage{times}
\usepackage[english]{babel}

\usepackage{ltexpprt}
\usepackage{algorithmic}
\usepackage{algorithm}
\usepackage{amsmath}
\usepackage{amssymb}
\usepackage{verbatim}
\usepackage{multirow}
\usepackage{booktabs}
\usepackage{url}
\usepackage{cite}
\usepackage{xspace}
\usepackage{array}
\usepackage{dcolumn}
\usepackage{graphicx}
\usepackage[tight,center]{subfigure}
\usepackage{color}
\usepackage{makeidx}
\usepackage{verbatim}
\usepackage{caption}
\usepackage{float}
\usepackage{enumitem}
\usepackage{pifont}
\usepackage{enumitem} 
\usepackage{calc} 

\usepackage{tikz}
\usetikzlibrary{snakes,arrows,shapes,positioning,fit,calc,patterns}
\usepackage{pgfplots}

\newcommand{\otoprule }{\midrule[\heavyrulewidth]}

\newcommand{\ourmaintitle}{Linear-time Detection of Non-linear Changes\\ in Massively High Dimensional Time Series}
\newcommand{\ourtitle}{\ourmaintitle}
\newcommand{\oururl}{\url{http://eda.mmci.uni-saarland.de/light/}}

\newcommand{\codeurl}{\oururl}

\newcommand{\ourmethod}{\textsc{Light}\xspace}
\newcommand{\spll}{\textsc{spll}\xspace}

\newcommand{\pind}{\textsc{pind}\xspace}
\newcommand{\lind}{\textsc{Light}$_{\mathit{ind}}$\xspace}

\newcommand{\lnf}{\textsc{Light}$_{\mathit{nf}}$\xspace}
\newcommand{\lnp}{\textsc{Light}$_{\mathit{np}}$\xspace}
\newcommand{\rsif}{\textsc{rsif}\xspace}

\newcommand{\pca}{\textsc{pca}\xspace}
\newcommand{\svd}{\textsc{svd}\xspace}

\newcommand{\kl}{\textsc{KL}\xspace}

\newcommand{\Ab}{A}

\newcommand{\graph}{\mathcal{G}}
\newcommand{\node}{\mathcal{V}}
\newcommand{\edge}{\mathcal{E}}
\newcommand{\Xb}{\mathbf{X}}
\newcommand{\X}{X}
\newcommand{\Yb}{\mathbf{Y}}

\newcommand{\wref}{\mathit{\mathcal{W}_{ref}}}
\newcommand{\wtest}{\mathit{\mathcal{W}_{test}}}
\newcommand{\wreft}{\mathit{\mathcal{W}_{ref}^{trans}}}
\newcommand{\wtestt}{\mathit{\mathcal{W}_{test}^{trans}}}
\newcommand{\sref}{\mathit{\mathcal{S}_{ref}}}
\newcommand{\stest}{\mathit{\mathcal{S}_{test}}}
\newcommand{\Space}{\mathbb{S}}
\newcommand{\score}{\mathit{score}}
\newcommand{\degr}{\mathit{deg}}
\newcommand{\corr}{\mathit{corr}}
\newcommand{\Size}{M}
\newcommand{\size}{m}
\newcommand{\dima}{n}
\newcommand{\diff}{\mathit{div}}

\newcommand{\maxv}{\mathit{V}}
\newcommand{\minv}{\mathit{v}}

\newcommand{\proofApx}{
\begin{proof}
\ifapx 
 We postpone the proof to Appendix~\ref{sec:proofs}.
\else 
 We postpone the proof to the online Appendix.
\fi 
\end{proof}
}

\makeatletter
\renewcommand*{\@fnsymbol}[1]{\ensuremath{\ifcase#1\or   \circ\or \bullet\or *\or \ddagger\or
   \mathsection\or \mathparagraph\or \|\or **\or \dagger\dagger
   \or \ddagger\ddagger \else\@ctrerr\fi}}
\makeatother

\pgfdeclarelayer{background}
\pgfdeclarelayer{foreground}
\pgfsetlayers{background,main,foreground}

\tikzstyle{block} = [rounded corners, draw=blue!70, fill=white, text width=3.3cm, minimum height=4em]
\tikzstyle{bgblock} = [rounded corners, draw=blue!70, thick, fill=blue!10, text width=3.3cm, minimum height=4em]
\tikzstyle{line} = [draw, -latex', thick,blue!70]

\definecolor{yafaxiscolor}{rgb}{0.3, 0.3, 0.3}

\definecolor{yafcolor1}{rgb}{0.4, 0.165, 0.553}
\definecolor{yafcolor2}{rgb}{0.949, 0.482, 0.216}
\definecolor{yafcolor3}{rgb}{0.47, 0.549, 0.306}
\definecolor{yafcolor4}{rgb}{0.925, 0.165, 0.224}
\definecolor{yafcolor5}{rgb}{0.141, 0.345, 0.643}
\definecolor{yafcolor6}{rgb}{0.965, 0.933, 0.267}
\definecolor{yafcolor7}{rgb}{0.627, 0.118, 0.165}
\definecolor{yafcolor8}{rgb}{0.878, 0.475, 0.686}

\newlength{\yafaxispad}
\newlength{\yaftlpad}
\newlength{\yaflabelpad}
\newlength{\yafaxiswidth}
\newlength{\yafticklen}
\makeatletter
\def\pgfplots@drawtickgridlines@INSTALLCLIP@onorientedsurf#1{}
\makeatother

\newcommand{\yafdrawaxes}[4]{
	\pgfplotstransformcoordinatex{#1}\let\xmincoord=\pgfmathresult 
	\pgfplotstransformcoordinatex{#2}\let\xmaxcoord=\pgfmathresult 
	\pgfplotstransformcoordinatey{#3}\let\ymincoord=\pgfmathresult 
	\pgfplotstransformcoordinatey{#4}\let\ymaxcoord=\pgfmathresult 
	\pgfsetlinewidth{\yafaxiswidth} 
	\pgfsetcolor{yafaxiscolor}
	\pgfpathmoveto{\pgfpointadd{\pgfpointadd{\pgfplotspointrelaxisxy{0}{0}}{\pgfqpointxy{\xmincoord}{0}}}{\pgfqpoint{-0.5\yafaxiswidth}{\yafaxispad}}}
	\pgfpathlineto{\pgfpointadd{\pgfpointadd{\pgfplotspointrelaxisxy{0}{0}}{\pgfqpointxy{\xmaxcoord}{0}}}{\pgfqpoint{0.5\yafaxiswidth}{\yafaxispad}}}
	\pgfpathmoveto{\pgfpointadd{\pgfpointadd{\pgfplotspointrelaxisxy{0}{0}}{\pgfqpointxy{0}{\ymincoord}}}{\pgfqpoint{\yafaxispad}{-0.5\yafaxiswidth}}}
	\pgfpathlineto{\pgfpointadd{\pgfpointadd{\pgfplotspointrelaxisxy{0}{0}}{\pgfqpointxy{0}{\ymaxcoord}}}{\pgfqpoint{\yafaxispad}{0.5\yafaxiswidth}}}
	\pgfusepath{stroke}
}

\newcommand{\yafdrawYaxis}[2]{
	\pgfplotstransformcoordinatey{#1}\let\ymincoord=\pgfmathresult 
	\pgfplotstransformcoordinatey{#2}\let\ymaxcoord=\pgfmathresult 
	\pgfsetlinewidth{\yafaxiswidth} 
	\pgfsetcolor{yafaxiscolor}
	\pgfpathmoveto{\pgfpointadd{\pgfpointadd{\pgfplotspointrelaxisxy{0}{0}}{\pgfqpointxy{0}{\ymincoord}}}{\pgfqpoint{\yafaxispad}{-0.5\yafaxiswidth}}}
	\pgfpathlineto{\pgfpointadd{\pgfpointadd{\pgfplotspointrelaxisxy{0}{0}}{\pgfqpointxy{0}{\ymaxcoord}}}{\pgfqpoint{\yafaxispad}{0.5\yafaxiswidth}}}
	\pgfusepath{stroke}
}

\newcommand{\yafdrawaxisLimits}[4]{
	\pgfplotstransformcoordinatex{#1}\let\xmincoord=\pgfmathresult
	\pgfplotstransformcoordinatex{#2}\let\xmaxcoord=\pgfmathresult
	\pgfplotstransformcoordinatey{#3}\let\ymincoord=\pgfmathresult 
	\pgfplotstransformcoordinatey{#4}\let\ymaxcoord=\pgfmathresult 
	\pgfsetlinewidth{\yafaxiswidth} 
	\pgfsetcolor{yafaxiscolor}
	\pgfpathmoveto{
		\pgfpointadd{
			\pgfpointadd{
				\pgfplotspointrelaxisxy{0}{0}}{
				\pgfqpointxy{\xmincoord}{0}
			}
		}{
			\pgfqpoint{
				-0.5\yafaxiswidth}{
				\yafaxispad
			}
		}
	}
	\pgfpathlineto{
		\pgfpointadd{
			\pgfpointadd{
				\pgfplotspointrelaxisxy{0}{0}
			}{
				\pgfqpointxy{
					\xmaxcoord
				}{0}
			}
		}{
			\pgfqpoint{
				25.5\yafaxiswidth
			}{
				\yafaxispad
			}
		}
	}
	\pgfpathmoveto{\pgfpointadd{\pgfpointadd{\pgfplotspointrelaxisxy{0}{0}}{\pgfqpointxy{0}{\ymincoord}}}{\pgfqpoint{\yafaxispad}{-0.5\yafaxiswidth}}}
	\pgfpathlineto{\pgfpointadd{\pgfpointadd{\pgfplotspointrelaxisxy{0}{0}}{\pgfqpointxy{0}{\ymaxcoord}}}{\pgfqpoint{\yafaxispad}{0.5\yafaxiswidth}}}
	\pgfusepath{stroke}
}

\pgfplotscreateplotcyclelist{yaf}{%
{yafcolor1,mark options={scale=0.75},mark=o}, 
{yafcolor2,mark options={scale=0.75},mark=square},
{yafcolor3,mark options={scale=0.75},mark=triangle},
{yafcolor4,mark options={scale=0.75},mark=o},
{yafcolor5,mark options={scale=0.75},mark=o},
{yafcolor6,mark options={scale=0.75},mark=o},
{yafcolor7,mark options={scale=0.75},mark=o},
{yafcolor8,mark options={scale=0.75},mark=o}} 

\pgfplotscreateplotcyclelist{yaf fill}{%
{fill=yafcolor1,draw=none}, 
{fill=yafcolor2,draw=none},
{fill=yafcolor3,draw=none},
{fill=yafcolor4,draw=none},
{fill=yafcolor5,draw=none},
{fill=yafcolor6,draw=none},
{fill=yafcolor7,draw=none},
{fill=yafcolor8,draw=none}} 

\pgfplotsset{jv ybar/.style={
   ybar, 
   cycle list name=yaf fill,
   xtick = \empty,
   every extra x tick/.style={major tick length=0pt,color=black}, 
   xmajorgrids = false,
}}

\pgfplotsset{jv line/.style={
   no markers,
   cycle list name=yaf,
   log ticks with fixed point,
   y tick label style = {/pgf/number format/set thousands separator = {\,}},
}}
\pgfplotsset{jv line ylog/.style={
   jv line,
}}

\pgfplotsset{axis y line=left, axis x line=bottom,
	tick align=outside,
	tickwidth=\yafticklen,
	clip = false,
    x axis line style= {-, line width = 0pt, color=black!0},
    y axis line style= {-, line width = 0pt, color=black!0},
    x tick style= {line width = \yafaxiswidth, color=yafaxiscolor, yshift = \yafaxispad},
    y tick style= {line width = \yafaxiswidth, color=yafaxiscolor, xshift = \yafaxispad},
    x tick label style = {font=\scriptsize, yshift = \yaftlpad},
    y tick label style = {font=\scriptsize, xshift = \yaftlpad},
    every axis y label/.style = {at = {(ticklabel cs:0.5)}, rotate=90, anchor=center, font=\scriptsize, yshift = -\yaflabelpad},
    every axis x label/.style = {at = {(ticklabel cs:0.5)}, anchor=center, font=\scriptsize, yshift = \yaflabelpad},
    x tick label style = {font=\scriptsize, yshift = 1pt},
    grid = major,
    major grid style  = {dash pattern = on 1pt off 3 pt},
	every axis plot post/.append style= {line width=\yafaxiswidth} ,
	legend cell align = left,
	legend style = {inner sep = 1pt, cells = {font=\scriptsize}},
	legend image code/.code={%
		\draw[mark repeat=2,mark phase=2,#1] 
		plot coordinates { (0cm,0cm) (0.15cm,0cm) (0.3cm,0cm) };%
	} 
}

\begin{document}

\title{\ourtitle}

\author{
Hoang-Vu Nguyen\thanks{Max Planck Institute for Informatics and Saarland University, Germany. Email: \texttt{\{hnguyen,jilles\}@mpi-inf.mpg.de}} \hspace{2.0cm}
Jilles Vreeken\footnotemark[1]
}

\date{}

\maketitle

\begin{abstract}
\small\baselineskip=9pt%
Change detection in multivariate time series has applications in many domains, including health care and network monitoring. A common approach to detect changes is to compare the divergence between the distributions of a reference window and a test window. When the number of dimensions is very large, however, the na\"ive approach has both quality and efficiency issues: to ensure robustness the window size needs to be large, which not only leads to missed alarms but also increases runtime. 

To this end, we propose \ourmethod, a linear-time algorithm for robustly detecting non-linear changes in massively high dimensional time series. Importantly, \ourmethod provides high flexibility in choosing the window size, allowing the domain expert to fit the level of details required. To do such, we 1)~perform scalable \pca to reduce dimensionality, 2)~perform scalable factorization of the joint distribution, and 3)~scalably compute divergences between these lower dimensional distributions. Extensive empirical evaluation on both synthetic and real-world data show that \ourmethod outperforms state of the art with up to 100\% improvement in both quality and efficiency.

\end{abstract}

\section{Introduction} \label{sec:intro}

Change detection in time series is an important task in data mining. It has applications in many areas, including health care and network monitoring~\cite{charu:change,kifer:change,liu:ulsif}. In principle, it is concerned with detecting time points at which important statistical properties of the time series change. To do so, a common approach is to compare the divergence between data distributions of a reference window and a test window. Traditionally, this works well for univariate time series~\cite{kifer:change,bifet:change,takeuchi:change}. Time series nowadays, however, are massively high dimensional. For instance, time series from human physical activities have more than 5\,000 dimensions~\cite{reyes:human}, while those from online URLs have more than 50\,000 dimensions~\cite{justin:url}. In these very high dimensional settings, existing work has both quality and efficiency issues.

In particular, techniques working directly with distributions of \textit{all} dimensions, such as~\cite{desobry:change,harchaoui:change,kawa:kliep}, are prone to the curse of dimensionality. Na\"ively, this issue seems to be resolved by using a large window size. A large window, however, in general increases runtime, causes high delay, and misses alarms~\cite{charu:change}. More recent work~\cite{kuncheva:change,qahtan:change} alleviates the issue by principal component analysis (\pca)~\cite{pca}. Using \pca they map data in reference and test windows to a lower dimensional space on which divergence assessment is carried out per \textit{individual} eigenvectors. While this in theory can avoid high dimensionality, it has three pitfalls. First, \pca is cubic in the number of dimensions. Second, to ensure the stability of covariance matrices that \pca uses, these methods need a large window size~\cite{ledoit:cov}. Third, by assuming statistical independence in the \pca space, they may miss complex changes over correlations of dimensions. Thus, detecting changes in massively high dimensional time series still is an open problem.

In this paper, we aim at addressing this issue. We do so by proposing \ourmethod, for \textbf{li}near-time change detection in hi\textbf{gh} dimensional \textbf{t}ime series. In short, \ourmethod overcomes the drawbacks of existing techniques by a three-step approach. First, we perform scalable \pca mapping of the two windows using matrix sampling~\cite{drineas:svd}. With matrix sampling, we allow for medium to small window sizes. After \pca mapping, we obtain data embedded in a much lower-dimensional space. To further increase robustness of divergence assessment and flexibility in choosing the window size, we factorize the joint distribution in the transformed space to lower dimensional distributions. Finally, we propose a divergence measure that computes divergence score using these distributions. Importantly, our measure has the non-negativity property of well-known divergence measures. Further, it allows non-parametric computation on empirical data in closed form. It also permits incremental calculation and hence is suited to the setting of time series. Our analysis and extensive experiments on both synthetic and real-world high dimensional time series show that \ourmethod scales linearly in the number of dimensions while achieving high quality.

The road map is as follow. In Sect.~\ref{sec:frame}, we give an overview of \ourmethod. In Sect.~\ref{sec:ins}, we provide its details and analyze its scalability. In Sect.~\ref{sec:rl}, we review related work. In Sect.~\ref{sec:exp}, we report empirical study. We round up with a discussion in Sect.~\ref{sec:dis} and conclude in Sect.~\ref{sec:con}. For readability, we put all proofs in the appendix.

\section{Overview} \label{sec:frame}

Consider an $\dima$-dimensional real-valued time series $\{\Xb(1), \Xb(2), \ldots, \Xb(\Size)\}$. At each time instant $t \in [1, \Size]$ the sample $\Xb(t)$ consists of $\dima$ values $\{\X_1(t), \ldots, \X_\dima(t)\}$ where $\X_i(t)$ corresponds to dimension $\X_i$ ($i \in [1, \dima]$). We assume that $\X_i(t) \in [\minv_i, \maxv_i]$ for all $t \in [1, \Size]$. Note that when the dimension scales are very different, we can normalize to bring them to comparable domains. Here we use different scales to provide more flexibility.

An overview of \ourmethod is given in Algorithm~\ref{algo:cd}. It works as follows. First, we form a reference window $\wref$ with size $|\wref| = \size$ (Line~2). Next, we transform $\wref$ to $\wreft$ of lower dimensionality and extract structure $\sref$ from $\wreft$ (Line~3). We will explain the rationale of this step shortly. For each new sample $\Xb(t)$ of the time series, we form a test window $\wtest = \{\Xb(t - \size + 1), \ldots, \Xb(t)\}$ (Lines~4 and~7). Next, we apply to $\wtest$ the transformation that we perform on $\wref$ to obtain $\wtestt$; we also extract structure $\stest$ based on $\sref$. Both steps are described in Lines~5 and~7. We define the change score as the divergence between $(\wreft, \sref)$ and $(\wtestt, \stest)$ (Line~8). When a change indeed happens (Line~9), we report it (Line~10) and restart the process. Note that in this work, we do not focus on handling time-dependent information, e.g.\ auto-correlation. Such information can be captured by patching multiple windows as done in~\cite{kawa:kliep,liu:ulsif}.

In \ourmethod, we transform $\wref$ by mapping it to another space $\Space$ of $k \ll \size$ dimensions. We achieve this with scalable \pca using matrix sampling~\cite{drineas:svd}. The purpose of this step is multi-fold. First, in many domains the the default dimensions may not reflect the true physical dimensions~\cite{NonlinearBook}; hence a transformation is done to alleviate this issue. Second, when $\dima$ is very large we need to choose a large window size $\size$ to ensure the robustness of divergence computation. A large $\size$ however causes high delay and misses alarms~\cite{charu:change}. Hence, dimension reduction is necessary to ensure quality. Third, by reducing the number of dimensions and allowing a smaller window size we are able to boost efficiency.

The usual next step would be to perform divergence assessment on $\Space$. Working with $k$-dimensional joint distributions however may still expose us to the curse of dimensionality~\cite{nguyen:cjs}. For instance, when $\dima = 50\,000$ and $k = 100$ it is true that $k \ll \dima$; yet, to effectively process a joint distribution with $k = 100$ dimensions we may still require a moderate $\size$. To tackle this we propose to factorize the joint distribution of $\wreft$ to a combination of selected lower dimensional distributions, preserving non-linear structures, e.g.\ non-linear correlations among $k$ dimensions of $\Space$. The set of lower-dimensional distributions makes up structure $\sref$ in our algorithm. The change score at time instant $t$ is computed based on these low dimensional distributions. Hence, we achieve three goals at the same time: (1) more flexibility in setting $\size$, (2) maintaining robustness of divergence assessment, and (3) detecting changes in complex non-linear structures. Note that the linear transformation with scalable \pca and the construction of $\sref$ could be perceived as compression of the data preserving first-order and higher-order dependencies, respectively.

For divergence assessment, we propose a new divergence measure that judiciously combines component lower dimensional distributions to produce the change score for the original distributions. Our measure has the important non-negativity property of popular divergence measures. In addition, it allows computation on empirical data in closed form. It also facilitates incremental computation and hence is suitable for time series application. Through the development of our measure, we show that when the dimensions in $\Space$ are indeed statistically independent, \ourmethod reduces to existing \pca-based change detection methods~\cite{kuncheva:change,qahtan:change}, which focus exclusively on linear relationships. Thus, \ourmethod is more general, i.e.\ it can uncover different types of change, be them linear or non-linear.

Having given an overview of \ourmethod, in the next section we provide its details and explain why it is a highly scalable solution for very high dimensional time series.

\begin{algorithm}[t]
\caption{\textsc{\ourmethod}}
\label{algo:cd}
\begin{algorithmic}[1]

\STATE Set $t_s = 0$

\STATE Set $\wref = \{\Xb(t_s + 1), \ldots, \Xb(t_s + \size)\}$

\STATE Set $[\wreft, \sref] = \textit{transformAndExtract}(\wref)$

\STATE Set $\wtest = \{\Xb(t_s + \size + 1), \ldots, \Xb(t_s + 2 \size)\}$
	
\STATE Set $[\wtestt, \stest] = \textit{apply}(\wtest, \wreft, \sref)$

\FOR{each new sample $\Xb(t)$ of the time series}
	\STATE Update $\wtest$, $\wtestt$, and $\stest$ by replacing $\Xb(t - \size)$ by $\Xb(t)$
	
	\STATE Set $\score = \diff(\wreft, \sref, \wtestt, \stest)$
	
	\IF{$\textit{change}(\score)$}
		\STATE Report a change at time instant $t$ and set $t_s = t$
		
		\STATE Repeat from step~2
	\ENDIF
\ENDFOR
\end{algorithmic}
\end{algorithm}

\section{Details of \ourmethod} \label{sec:ins}

In this section we provide the details of \ourmethod, namely the data transformation, the distribution factorization, the divergence measure, and the change score thresholding.


\subsection{Data Transformation}
\label{sec:ins:transform}


Following the previous section, we perform \pca on the reference window $\wref$ with $\size$ points and $\dima$ dimensions. For simplicity we use $\Ab$ to denote the data of $\wref$; it has $\size$ rows and $\dima$ columns. W.l.o.g., we assume that $\Ab$ is centered. The usual procedure would be to (1) solve the eigendecomposition problem on $\Ab^T \Ab$, and (2) use the top eigenvectors corresponding to the top eigenvalues to transform $\Ab$. This costs $O(\size \dima^2 + \dima^3)$. When $\dima$ is very large this complexity is prohibitive for large scale processing. Further, when $\dima$ is large we need to choose a large window size $\size$; otherwise $\Ab^T \Ab$ becomes very unstable making \pca results unreliable~\cite{ledoit:cov}.

We overcome this by matrix sampling. In short, matrix sampling is concerned with sampling rows/columns of matrices to reduce the computational cost of common procedures, e.g.\ matrix multiplication and \svd. These approximations come with guaranteed error bounds. Further, as the sampled submatrix has fewer columns the results are also more stable~\cite{drineas:svd}. Matrix sampling has been recently applied in the context of data mining~\cite{liberty:sketching}. Here, we employ the technique in~\cite{drineas:svd} as it allows us to avoid fixing a priori the number of dimensions $k$ to be kept. This method essentially performs approximate singular value decomposition (\svd). Recall that \svd finds matrices $U$, $\Sigma$, and $V$ such that $\Ab = U \Sigma V^T$. Here, $U \in \mathbb{R}^{\size \times \size}$ and $V \in \mathbb{R}^{\dima \times \dima}$ contain the eigenvectors of $\Ab \Ab^T$ and $\Ab^T \Ab$, respectively. $\Sigma = \mathit{\textbf{diag}}(\lambda_1, \ldots, \lambda_d)$ is a diagonal matrix of size $\size \times \dima$ where $d = \min\{\size, \dima\}$. The singular values $\lambda_1 \geq \ldots \geq \lambda_d$ of A are also the non negative square roots of the eigenvalues of both $\Ab \Ab^T$ and $\Ab^T \Ab$. Finding the exact solution of \svd, like \pca, costs $O(\size \dima^2 + \dima^3)$. With matrix sampling we obtain high quality approximate solution of \svd in much less time. The details are as follows.

We write the column $j$ of $\Ab$ as $\Ab^{(j)}$ for $j \in [1, \dima]$. First, we sample with replacement $c \ll \dima$ columns of $\Ab$ according to their relative variance $\frac{|\Ab^{(j)}|^2}{||\Ab||^2_F}$ where $|\Ab^{(j)}|$ is the squared norm of vector $\Ab^{(j)}$ and $||.||_F$ is the Frobenius norm. This forms a matrix $C \in \mathbb{R}^{\size \times c}$. Second, we perform \pca on $C^T C$ to extract its top $k \leq c$ eigenvectors $\{y_1, \ldots, y_k\}$ corresponding to its top eigenvalues $\alpha_1 \geq \ldots \geq \alpha_k$. The value of $k$ is determined by how much variance of $C$ to preserve; usually 90\% to 99\% suffices~\cite{NonlinearBook,qahtan:change}. Next, we form matrix $U_k \in \mathbb{R}^{m \times k}$ where each column $U^{(i)}_k = \frac{Cy_i}{\sqrt{\alpha_i}}$, and matrix $\Sigma_{k} = \mathit{\textbf{diag}}(\sqrt{\alpha_1}, \ldots, \sqrt{\alpha_k})$ of size $k \times k$. Let $V_k \in \mathbb{R}^{\dima \times k}$ be a matrix whose columns contain the top $k$ eigenvectors of $\Ab^T \Ab$. According to~\cite{drineas:svd}, $\Ab_k \approx U_k \Sigma_k V_k^T$ where $\Ab_k$ is the best rank $k$ approximation of $\Ab$ w.r.t.\ both $||.||_F$ and $||.||_2$ (spectral norm). By definition, the \pca-transformed data $\wreft$ is given by $\Ab V_k = \Ab_k V_k = U_k \Sigma_k$. We also need to compute $V_k$ to transform the test window $\wtest$ later. It holds that $V_k = \Ab^T (U_k \Sigma_k) = \Ab^T (\Ab V_k)$. The original algorithm is stochastic. We can make it deterministic by picking $c$ columns of $\Ab$ with largest relative variance; to break ties we use a canonical order. 

The cost of approximating top $k$ eigenvectors $V_k$ of $\Ab^T \Ab$ is $O(\size c^2 + c^3)$. The cost of computing $\Ab V_k$ is $O(\size k^2)$. The cost of computing $V_k$ is $O(\size \dima k)$. So the total complexity is $O(\size c^2 + c^3 + \size k^2 + \size \dima k)$ with $k \leq c \ll \dima$, which can be simplified to $O(\size c^2 + \size \dima k)$ as we choose $c \leq \size$. With reasonable $c$ and small $\size$ the accuracy is increased~\cite{drineas:svd}, i.e.\ small window size is favored. The complexity of this sampling algorithm is on par with other sampling methods, such as~\cite{sarlos:sampling,liberty:sketching}. While these techniques require us to provide the final dimensionality $k$, our method permits to adapt $k$ to the current reference window $\Ab = \wref$. Overall, by matrix sampling we are able to boost efficiency and gain stability, as we perform \pca on $C^T C$ where $c \ll \dima$ instead of $\Ab^T \Ab$. In other words, besides guaranteed error bounds the approximate solutions will tend to be more stable to high dimensionality.

\subsection{Distribution Factorization}
\label{sec:ins:factor}

To further increase robustness of divergence assessment and flexibility in choosing the window size, we factorize joint distribution in the transformed space $\Space$ to low dimensional distributions.
We accomplish this with graphical modeling~\cite{wainwright:graph,graphbook}, a powerful approach for this purpose. We denote the $k$ dimensions of $\Space$ as $\{Y_1, \ldots, Y_k\}$. Assume that we have a graph $\graph = \{\node, \edge\}$ where $\node = \{Y_1, \ldots, Y_k\}$; two dimensions $Y_i$ and $Y_j$ not connected by an edge are regarded as conditionally independent given all other dimensions $\node \setminus \{Y_i, Y_j\}$. Given such a graph $\graph$, one can factorize the joint distribution $p(Y_1, \ldots, Y_k)$ of $\wreft$ by first finding special structures (e.g.\ cliques or connected components) of $\graph$, and then using the distributions of these structures to estimate $p(Y_1, \ldots, Y_k)$. Of course, the more complex the structures the better the estimation. Nevertheless, complex structures tend to contain many dimensions, causing quality and efficiency issues for divergence computation. We achieve a more balanced solution by using the edges of $\graph$, which is also a good alternative to factorize $p(Y_1, \ldots, Y_k)$~\cite{wainwright:graph,graphbook}. Under this model
$$p(Y_1, \ldots, Y_k) = \frac{\prod\limits_{(Y_i, Y_j) \in \edge} p(Y_i, Y_j)}{\prod\limits_{Y \in \node} p(Y)^{\degr(Y) - 1}}$$
where $\degr(Y)$ is the degree of $Y$ in $\graph$. One can see that the joint distribution is now represented by 1-D and 2-D distributions, further easing the `pressure' on picking window size $\size$. So far, we assume that $\graph$ is available. Now we describe how to obtain it from $\wreft$.

Our solution consists of three steps: 1)~computing pairwise correlation score of $Y_i$ and $Y_j$ with $i \neq j$ -- the higher the score the more correlated, 2)~initializing $\graph$ containing all dimensions and having an edge between every two dimensions -- the weight of each edge is their correlation score, and 3)~simplifying $\graph$ to one of its maximum spanning tree. Note that the simplified $\graph$ has $(k-1)$ edges.

Recall that our goal is to detect non-linear changes. To this end, for the first step we choose the quadratic measure of dependency~\cite{seth:measure} as it captures non-linear correlations, is non-parametric, and permits computation on empirical data in closed form. In short, under this measure the correlation between $Y_i$ and $Y_j$ is given by
$$\corr(Y_i, Y_j) = \int \int \left(P(y_i, y_j) - P(y_i) P(y_j)\right)^2 dy_i dy_j$$
where $P(.)$ denotes cumulative distribution function (cdf).
Computing this measure for all dimension pairs takes $O(\size^2 k^2)$. To boost efficiency, similar to~\cite{nguyen:4s} we apply AMS Sketch~\cite{alon:ams}. In short, to compute $\corr(Y_i, Y_j)$ we pre-compute the sketches of both $Y_i$ and $Y_j$ by projecting their realizations onto multiple random vectors $u \in \{-1, +1\}^{\size}$. We then utilize the sketch values to estimate $\corr(Y_i, Y_j)$. The estimate is unbiased and its error is bounded~\cite{alon:ams}. The more random vectors used the better the estimate. We find that using $O(k)$ vectors suffices. The time complexity of computing pairwise correlation scores is thus reduced to $O(\size k^2)$~\cite{nguyen:4s}. To make our method deterministic we generate the vectors in advance and reuse whenever needed.

For the third step, as $\graph$ initially is dense we employ the method proposed in~\cite{fredman:mst}. It is deterministic and costs $O(k^2)$. The outcome of this step is the set of 1-D and 2-D distributions taken from the maximum spanning tree. These distributions also constitute the structure $\sref$ of $\wreft$.

The overall complexity of distribution factorization is $O(\size k^2)$, i.e.\ linear in $\size$ and independent of $\dima$.

\subsection{Divergence Computation}
\label{sec:ins:diver}

For each test window $\wtest$ we first map it to $\Space$ using $V_k$ (see Section~\ref{sec:ins:transform}). Then, we use $\sref$ to extract $\stest$, i.e.\ the set of 1-D and 2-D distributions of $\wtest$ corresponding to those in $\sref$. Our goal now is to compute the divergence score
$$\score = \diff\left(p(Y_1, \ldots, Y_k)\; ||\; q(Y_1, \ldots, Y_k)\right)$$
where $p(Y_1, \ldots, Y_k)$ and $q(Y_1, \ldots, Y_k)$ are the joint distributions of $\wref$ and $\wtest$, respectively. One important question to address here is: How to do this using two sets of distributions $\sref$ and $\stest$? We answer this question based on the following observations.

\begin{lemma} \label{lem:klfactor}
Let $\kl\left(p(.)\; ||\; q(.)\right)$ be the Kullback-Leibler divergence between $p(.)$ and $q(.)$. Using $\sref$ and $\stest$ we have $\kl\left(p(Y_1, \ldots, Y_k)\; ||\; q(Y_1, \ldots, Y_k)\right) =$
\begin{align*}
&\textstyle\sum\limits_{(Y_i, Y_j) \in \edge} \kl\left(p(Y_i, Y_j)\; ||\; q(Y_i, Y_j)\right)\\
&\textstyle- \sum\limits_{Y \in \node: \degr(Y) > 1} (\degr(Y)-1) \kl\left(p(Y)\; ||\; q(Y)\right)
\end{align*}
\end{lemma}

\proofApx

Lemma~\ref{lem:klfactor} tells us that $\score$ w.r.t.\ \kl measure is equal to the sum of divergence scores of 2-D distributions (called 2-D divergence scores) offset by those of 1-D distributions (called 1-D divergence scores). Here, $\kl\left(p(Y_i, Y_j)\; ||\; q(Y_i, Y_j)\right)$ stands for the magnitude of changes in $Y_i$, $Y_j$, and their dependency. Thus, the sum of 2-D divergence scores stands for the magnitude of changes in the involved dimensions and their dependencies. The subtraction by 1-D divergence scores is to reduce the impact of dimensions contributing to more than one 2-D score term.

Though \kl divergence features a nice computation of $\score$ based on 1-D and 2-D distributions in $\sref$ and $\stest$, these distributions need to be estimated, e.g.\ parametrically or by kernel density estimation~\cite{scott:density}. Here, we aim at a purely non-parametric approach to maintain the exploratory nature of \ourmethod. Thus, we propose to \textit{generalize} the result in Lemma~\ref{lem:klfactor} to divergence measures other than \kl. In particular, for any measure $\diff$ we propose to set $\score$ to
\begin{align*}
&\textstyle\delta \times \left(\sum\limits_{(Y_i, Y_j) \in \edge} \diff\left(p(Y_i, Y_j)\; ||\; q(Y_i, Y_j)\right)\right)\\
&\textstyle- \sum\limits_{Y \in \node: \degr(Y) > 1} (\degr(Y)-1) \diff\left(p(Y)\; ||\; q(Y)\right)
\end{align*}
where $\delta$ is a regularization factor, which is to guarantee that $\score$ is non-negative. With our generalization we give way to applying other non-parametric instantiations for $\diff$, e.g.\ the one in~\cite{nguyen:ipd} with empirical computation in closed form, while still upholding the non-negativity property of \kl divergence. An additional benefit of $\delta$ is that it can make the influence of $Y$ with $\degr(Y) > 1$ on $\score$ more prominent, i.e.\ more impact is given to the dimensions correlated to multiple other dimensions.


Before introducing the setting of $\diff$ we show that \pca-based change detection methods~\cite{kuncheva:change,qahtan:change} -- using the previous two steps of our framework -- also generalize from Lemma~\ref{lem:klfactor}, yet in a more restrictive manner. In particular, these methods estimate $\score$ by $\sum\limits_{i=1}^k \diff\left(p(Y_i)\; ||\; q(Y_i)\right)$ or $\max\limits_{i \in [1, k]} \diff\left(p(Y_i)\; ||\; q(Y_i)\right)$; note that the two forms are similar. From Lemma~\ref{lem:klfactor} we see that if $p(Y_i, Y_j) = p(Y_i) p(Y_j)$ and $q(Y_i, Y_j) = q(Y_i) p(Y_j)$ for $(Y_i, Y_j) \in \edge$, then $\score$ under \kl is equal to $\sum\limits_{i=1}^k \kl\left(p(Y_i)\; ||\; q(Y_i)\right)$. Thus, \pca-based methods~\cite{kuncheva:change,qahtan:change} also generalize from \kl divergence; however, they impose two additional assumptions that $Y_i$ and $Y_j$ where $(Y_i, Y_j) \in \edge$ are statistically independent under both the data of $\wreft$ and $\wtestt$. We in turn do not impose these restrictions and can capture correlation in the \pca space $\Space$, i.e.\ we provide a more general solution.

\vspace{1.0em}
\noindent\textbf{Choosing $\diff$.}\
We use the quadratic measure of distribution divergence~\cite{nguyen:ipd}. It is purely non-parametric and its empirical computation is in closed form. Under this measure
$$\diff\left(p(Y)\; ||\; q(Y)\right) = \int \left(P(y) - Q(y)\right)^2 dy$$
and $\diff\left(p(Y_i, Y_j)\; ||\; q(Y_i, Y_j)\right)$ is defined similarly. We set $\textstyle\delta = 2 \sqrt{\sum\limits_{i=1}^{\dima} \max\{\minv_i^2, \maxv_i^2\}}$ following our below lemma.

\begin{lemma} \label{lem:score}
Setting $\diff$ to the measure in~\cite{nguyen:ipd} and $\delta$ as above, it holds that $\score \geq 0$ with equality iff $p(Y_1, \ldots, Y_k)$ and $q(Y_1, \ldots, Y_k)$ under the factorization model (cf., Section~\ref{sec:ins:factor}) are equal.
\end{lemma}

\proofApx

\noindent\textbf{Complexity analysis.}\
The cost of computing $\score$ for initial $\wref$ and $\wtest$ or after every change is $O(\size^2 k)$. The cost of computing divergence score for each new sample of the time series is $O(\size k)$; more details on this are in the Appendix~\ref{sec:incremental}. In case a large window size $\size$ is required, e.g.\ due to the application scenario, we can further boost the efficiency of our method by sampling with replacement the data of $\wreft$ and $\wtestt$. The details are in Appendix~\ref{sec:scalinglarge}.



\subsection{Change Score Thresholding}

We perform an adaptive thresholding scheme to decide when a change indeed happens. The scheme we consider is the Page-Hinkley test, which has been used in~\cite{qahtan:change}. Under this test we keep track of the past change scores corresponding to time instants without change. For a new time instant, we assess how much its change score deviates from these historical scores and raise the flag when the deviation is large (w.r.t.\ an easy-to-adapt threshold). More details are in~\cite{qahtan:change}.


\subsection{Summing Up} \label{sec:summing}

On a time series with $r$ changes, \ourmethod costs $O\left((\size c^2 + \size \dima k + \size^2 k) r + (\Size - r) \size k\right)$. In our experiment $\size = O(\dima)$, which simplifies the complexity to $O\left((c^2 + \dima k) \size r + (\Size - r) \size k\right)$. Thus, \ourmethod is linear in the window size $\size$, the original number of dimensions $\dima$, and the length $\Size$ of the time series. Being linear in $\dima$, it is suited to very high dimensional time series. Further, it can avoid large window size $\size$ while still achieving high quality, which is verified by our experiments in Section~\ref{sec:exp}.

\section{Related Work} \label{sec:rl}


\pca-based change detection is studied in~\cite{kuncheva:change,qahtan:change}. Kuncheva and Faithfull~\cite{kuncheva:change} propose to use eigenvectors with small eigenvalues for transformation. Qahtan et al.~\cite{qahtan:change} in turn show theoretically and empirically that eigenvectors corresponding to large eigenvalues instead are more relevant. Both methods apply \pca on original reference windows and have cubic runtime in the number of dimensions $\dima$. Further, they may miss complex changes due to the assumption that dimensions in the transformed space are independent, i.e.\ dimensions in the original space have linear correlations only.

Change detection based on estimating the ratio between distributions of reference and test windows is introduced in~\cite{kawa:kliep,liu:ulsif}. The main idea is to directly approximate this ratio by kernel models \textit{without} approximating the two distributions. This procedure implicitly assumes that data is uniformly distributed in the $\dima$-dimensional space. When $\dima$ is large, we need a large window size $\size$ to fill in this space to uphold this assumption. Computing the distribution ratio analytically costs $O(\size^3)$. For efficiency purposes the window size $\size$ must hence be kept small.

Song et al.~\cite{song:mixedg} propose a divergence test based on kernel density estimation for change detection. By performing density estimation on the original $\dima$-dimensional space, this test is susceptible to the curse of dimensionality~\cite{scott:density}. While one could apply this test after distribution factorization, proving the non-negativity of the resulting score is non-trivial. Other change detection techniques on multivariate time series~\cite{charu:change,desobry:change,harchaoui:change,Dasu:bootstrap} also work on the $\dima$-dimensional space, and hence, are also prone to the curse of dimensionality.

Our method in contrast alleviates the high dimensionality issue by scalable \pca mapping using matrix sampling. Furthermore, it does not make any assumption on data distributions nor that dimensions are linearly correlated. Lastly, it permits to set the windows size to match the level of details required by the domain expert.

\section{Experiments} \label{sec:exp}

In this section, we empirically evaluate \ourmethod. In particular, we study its effectiveness in detecting known change points on both synthetic and real-world data sets. We implemented \ourmethod in Java, and make our code available for research purposes.\!\footnote{\codeurl} All experiments were performed single-threaded on an Intel(R) Core(TM) i7-4600U CPU with 16GB RAM. We report wall-clock running times.

We compare to \pind~\cite{qahtan:change} and \spll~\cite{kuncheva:change}, two state of the art methods for \pca-based change detection. Both apply traditional \pca, assuming that dimensions in the \pca space are statistically independent. In addition, we consider \rsif~\cite{liu:ulsif}, which measures divergence scores by directly approximating density ratio of distributions. For each competitor, we optimize parameter settings according to their respective papers. \ourmethod has 4 parameters, namely the number of sampled dimensions $c$ which is used in scalable \pca mapping; the percentage of variance preserved which is used in scalable \pca mapping; the number of sketches $s_1$ and the number of average sketch values $s_2$ which are used in distribution factorization. Note that the last two parameters apply for any method using AMS Sketch~\cite{alon:ams}. The default setting for the parameters is: $c = 200$, percentage $= 90\%$, $s_1 = 50$, and $s_2 = 3$.

We experiment on both synthetic and real data sets. For the latter, we draw 7 data sets from the UCI Machine Learning Repository: Amazon, EMG Actions 1, EMG Actions 2, Human Activities, Human Postural Transitions, Sport Activities, and Youtube Video Games. All are high dimensional time series. As change points, we use existing class labels -- a common practice in change detection~\cite{kawa:kliep,song:mixedg}. Table~\ref{tab:datasets} summarizes the characteristics of these data sets.

\begin{table}[t]
\centering 
\begin{tabular}{lrr}
\toprule
{\bf Data} & {\bf $\Size$} & {\bf $\dima$}\\
\otoprule

Amazon & 30\,000 & 20\,000\\

EMG Actions 1 & 1\,800 & 3\,000\\

EMG Actions 2 & 3\,600 & 2\,500\\

Human Activities & 10\,299 & 561\\

Human Postural Transitions & 10\,929 & 561\\

Sport Activities & 9\,120 & 5\,625\\

Youtube Video Games & 120\,000 & 50\,000\\

\bottomrule
\end{tabular}
\caption{Characteristics of the real data sets. $\Size$ is the length of the time series and $\dima$ is its number of dimensions.} \label{tab:datasets} 
\end{table}

\subsection{Synthetic Data Generation}

Each $\dima$-dimensional time series we generate contains 100 segments, each having 2000 time steps. We create the change point between every two consecutive segments by varying either distributions or correlation patterns of some dimensions. This means that each time series has 99 change points. We evaluate change detection techniques based on how well they retrieve these known change points. In line with previous work~\cite{kuncheva:change,qahtan:change}, a change point detected by a method is considered to be a true positive if it is flagged correctly before $2\size$ points from the new distribution arrive. As performance metric, we use the F1 measure, which is defined as $\frac{2 \times \textit{precision} \times \textit{recall}}{\textit{precision} + \textit{recall}}$. Note that the higher the F1 score the better. Below we describe three different ways to create change points.

\noindent\textbf{Gaussian Distributions.}\ In this setting, vector $(\X_1, \ldots, \X_{2l})$ has multivariate Gaussian distribution ($l = \left\lfloor \frac{n}{3} \right\rfloor$). The mean vector and covariance matrix are initialized randomly. We consider three types of change: 1) change in mean vector, 2) change in individual variance terms, and 3) change in covariance terms. When creating segments, we switch among those three types of change in a round robin fashion. Each remaining dimension $\X_j$ where $j \in [2l + 1, \dima]$ in turn has its distribution fixed to $\textit{Gaussian}(0, 1)$.

\noindent\textbf{Linear Correlations.}\ We embed linear correlations in dimensions $\X_1, \ldots, \X_{2l}$ where $l = \left\lfloor \frac{n}{3} \right\rfloor$. To model correlation in each segment, we first generate $\Xb_{l \times 1} = \mathbf{A}_{l \times l} \times \mathbf{Z}_{l \times 1}$ where $Z_i \sim \textit{Gaussian}(0, 1)$ and $\mathbf{A}_{l \times l}$ is fixed with $a_{ij}$ initially drawn from $\textit{Uniform}[0, 1]$. Here, $\Xb_{l \times 1}$ and $\mathbf{Z}_{l \times 1}$ are two sets, each containing $l$ dimensions. We let $\{\X_1, \ldots, \X_l\} = \Xb_{l \times 1}$. Next, we generate $\mathbf{W}_{l \times 1} = \mathbf{B}_{l \times l} \times \Xb_{l \times 1}$ where $\mathbf{B}_{l \times l}$ is fixed with $b_{ij}$ initially drawn from $\textit{Uniform}[0, 0.5]$. Then, using a function $f$ we generate $\X_{i + l} = f(W_i) + e_i$ where $i \in [1, l]$, and $e_i \sim \textit{Gaussian}(0, \sigma)$; we fix $\sigma = 0.01$. We use two linear instantiations of $f$:
\begin{eqnarray} \nonumber
f_1(x) = 2x + 1, &\qquad& f_2(x) = \frac{x}{3} - 4. \nonumber
\end{eqnarray}
When creating time series segments, we alternatively pick $f_1$ and $f_2$. In this way, in every two consecutive segments the correlations between $\{\X_1, \ldots, \X_l\}$ and $\{\X_{l+1}, \ldots, \X_{2l}\}$ are different. Each dimension $\X_j$ where $j \in [2l + 1, \dima]$ is drawn from $\textit{Gaussian}(0, 1)$.

\noindent\textbf{Non-linear Correlations.}\ To embed non-linear correlations among dimensions $\X_1, \ldots, \X_{2l}$ where $l = \left\lfloor \frac{n}{3} \right\rfloor$, we follow the procedure as in linear correlations except for that we here use four non-linear and complex instantiations of $f$:
\begin{eqnarray} \nonumber
f_3(x) = x^2 - 2x\;, &\qquad& f_4(x) = x^3 + 3x + 1\;, \\ \nonumber
f_5(x) = \log(|x| + 1)\;, &\qquad& f_6(x) = \sin(2x)\;.
\end{eqnarray}
When creating time series segments, we  switch among $f_3$, $f_4$, $f_5$, and $f_6$ in a round robin fashion. Each dimension $\X_j$ where $j \in [2l + 1, \dima]$ is drawn from $\textit{Gaussian}(0, 1)$.

\subsection{Quantitative Results on Synthetic Data}

\begin{figure}[t]
\centering
\subfigure[Gaussian distributions: F1 vs.\ dimensionality]
{{\includegraphics[width=0.22\textwidth]{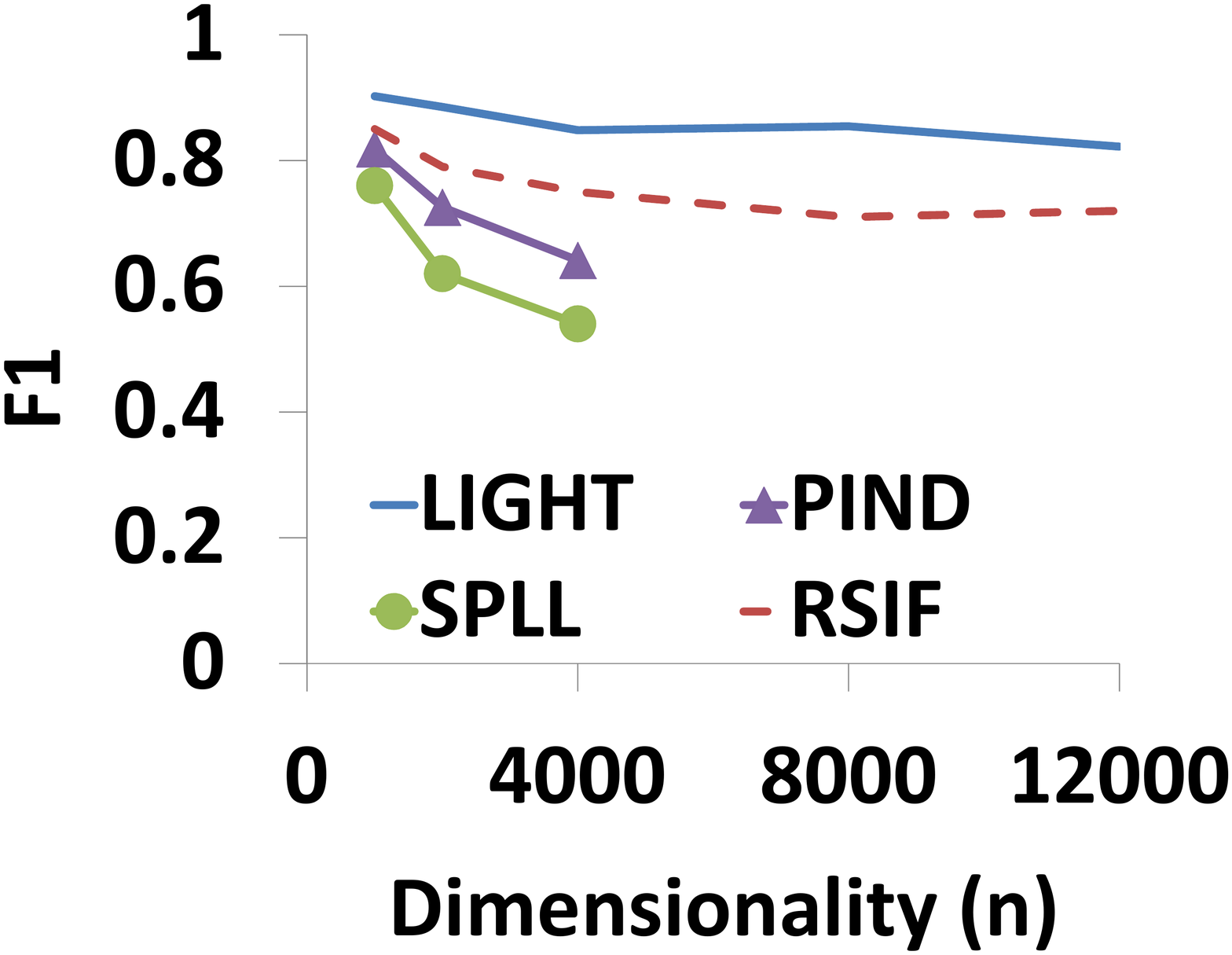}}\label{fig:acc_vs_dim_gaus}}
\subfigure[Gaussian distributions: Runtime vs.\ dimensionality]
{{\includegraphics[width=0.22\textwidth]{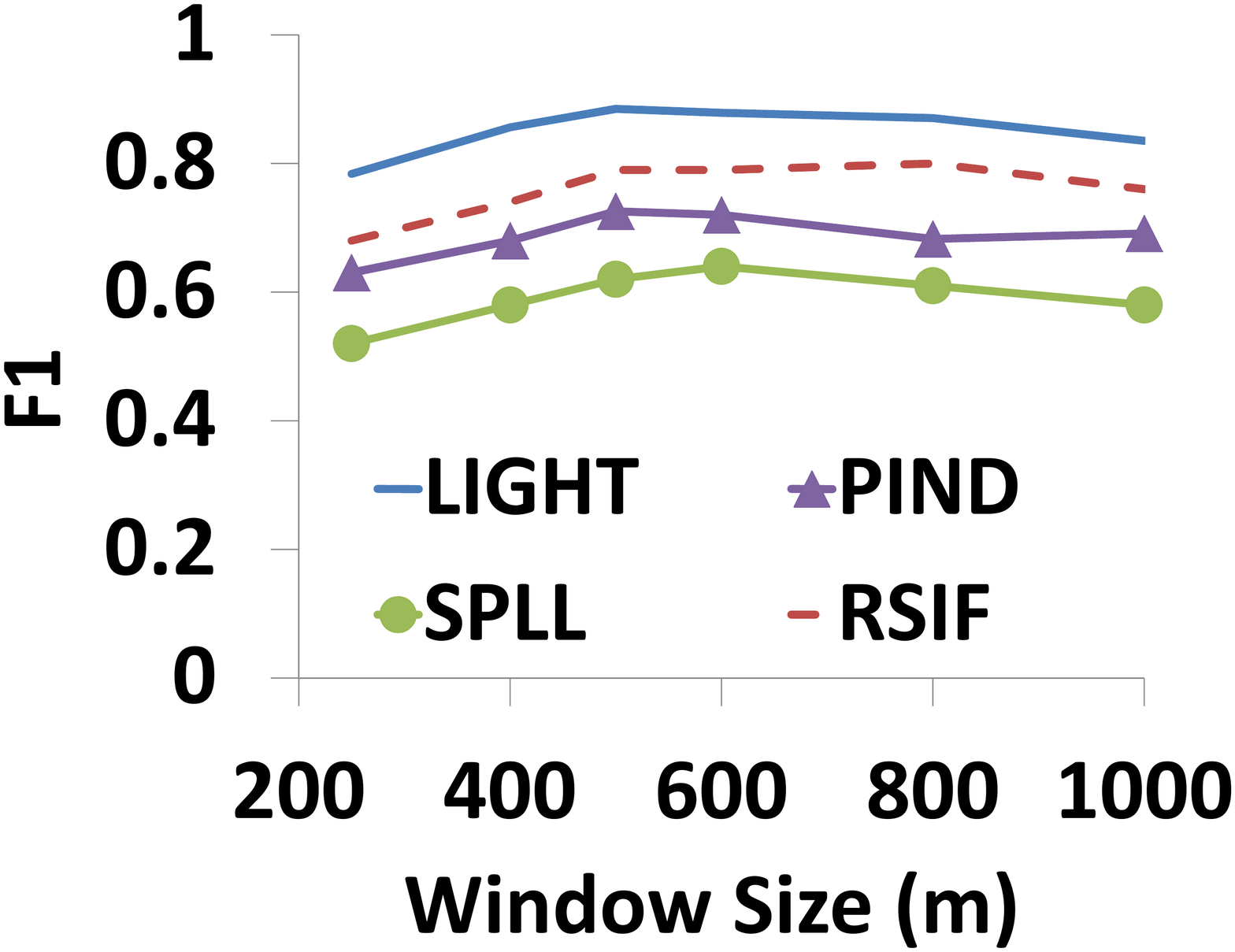}}\label{fig:acc_vs_win_gaus}}
\subfigure[Linear correlations: F1 vs.\ dimensionality]
{{\includegraphics[width=0.22\textwidth]{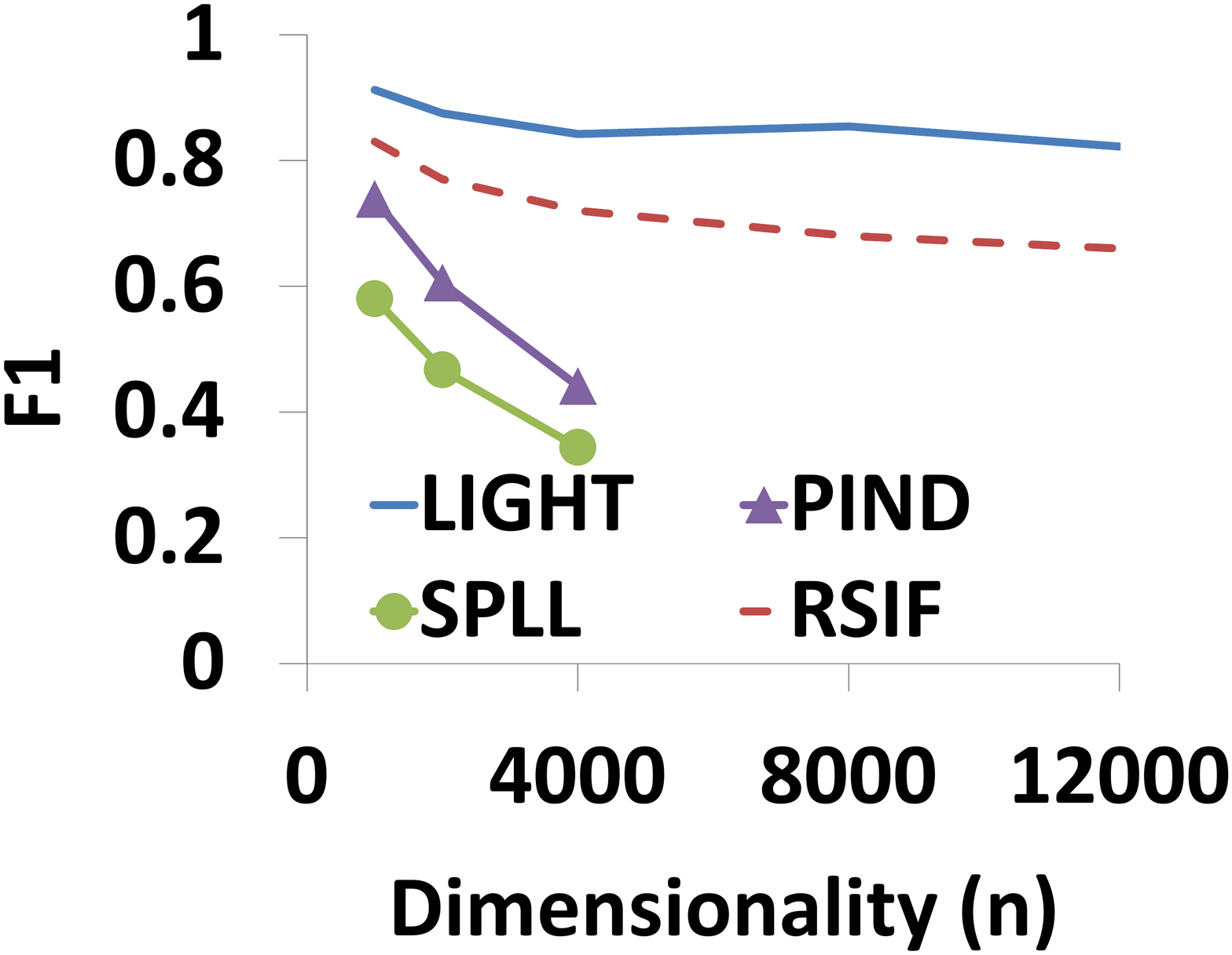}}\label{fig:acc_vs_dim_linear}}
\subfigure[Linear correlations: Runtime vs.\ dimensionality]
{{\includegraphics[width=0.22\textwidth]{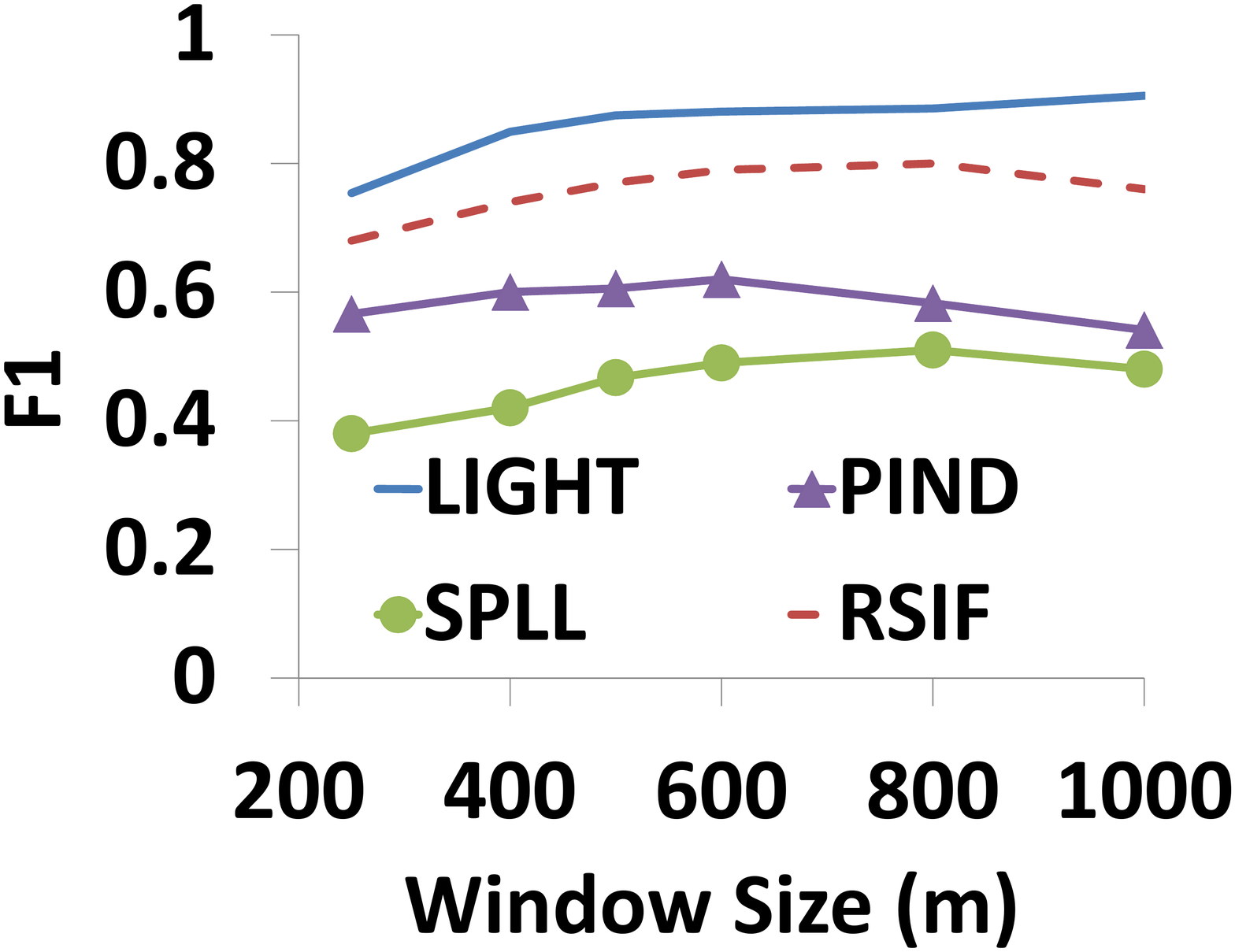}}\label{fig:acc_vs_win_linear}}
\subfigure[Non-linear correlations: F1 vs.\ dimensionality]
{{\includegraphics[width=0.22\textwidth]{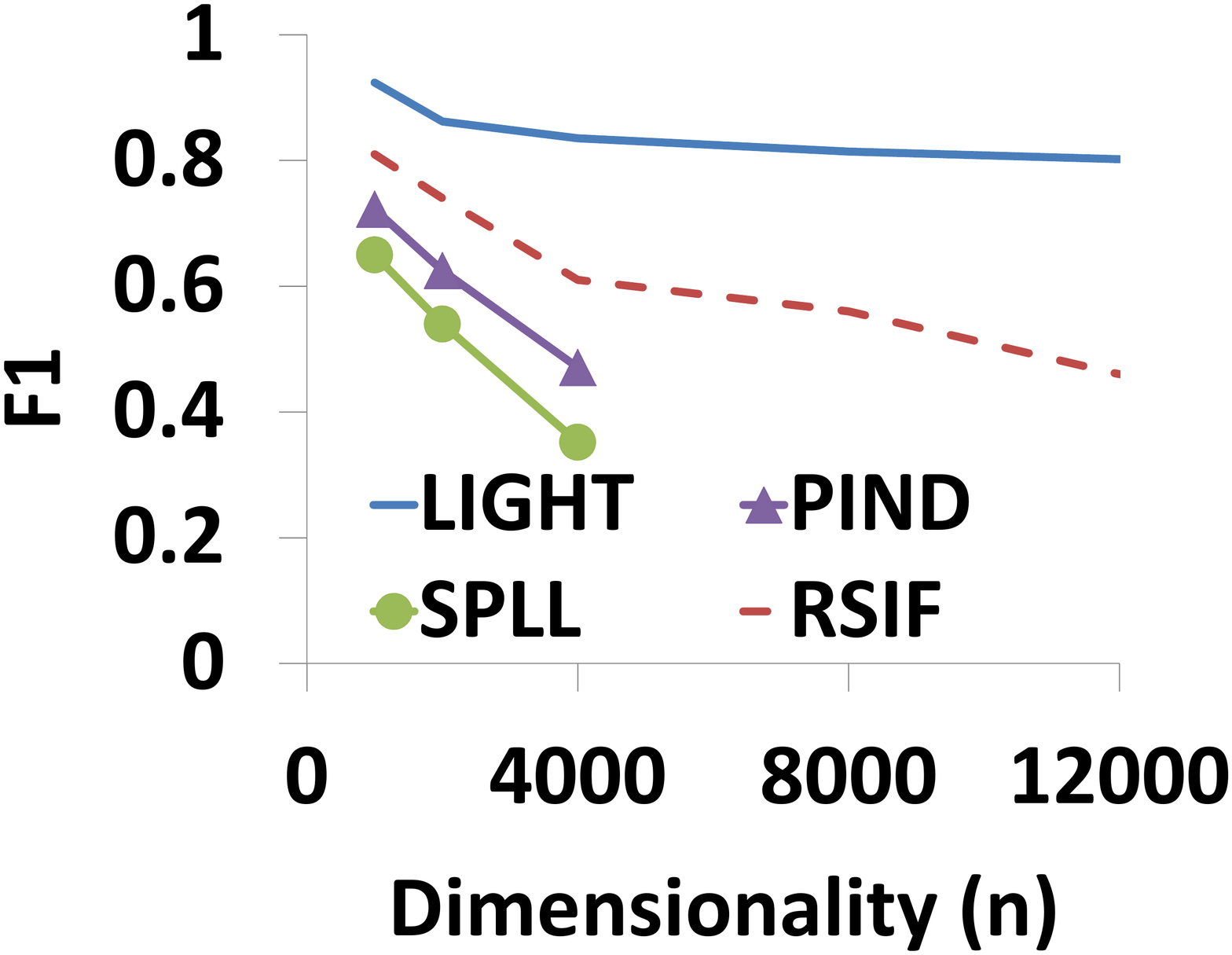}}\label{fig:acc_vs_dim_complex}}
\subfigure[Non-linear correlations: Runtime vs.\ dimensionality]
{{\includegraphics[width=0.22\textwidth]{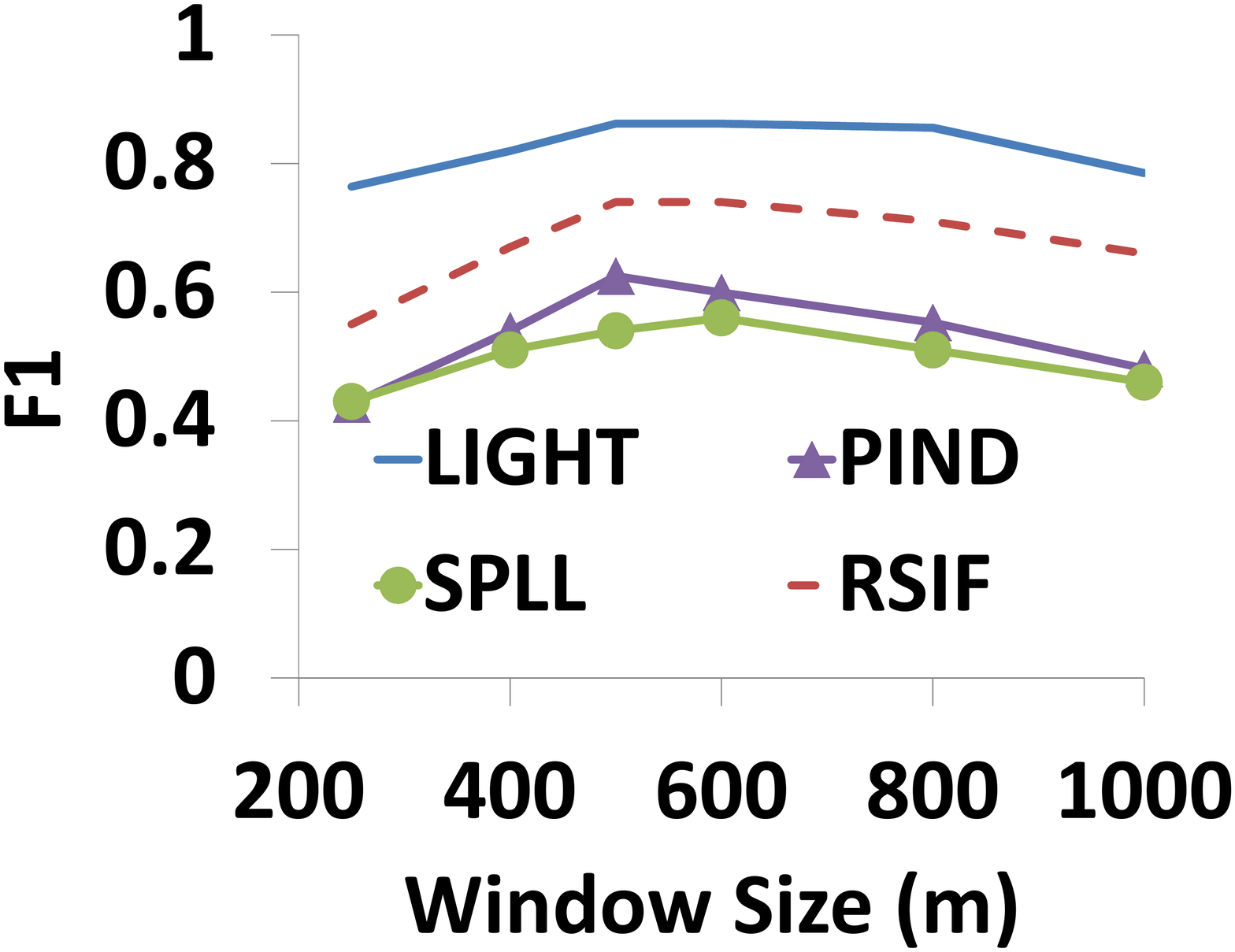}}\label{fig:acc_vs_win_complex}}
\caption{[Higher is better] Comparison with competitors: F1 scores on synthetic data sets. Overall, \ourmethod yields the best quality across different types of change, values of dimensionality $\dima$, and window sizes $\size$.}
\end{figure}

We assess quality of the methods tested under different values of dimensionality $\dima$ and window sizes $\size$. For quality against $\size$ we fix $\dima = 2000$. For quality against $\dima$ we fix $\size = 500$.
The results are in Figures~\ref{fig:acc_vs_dim_gaus}---\ref{fig:acc_vs_win_complex}.

We find that \ourmethod consistently achieves the best performance across different values of $\dima$ and $\size$, and different types of change -- from linear to highly non-linear. Its quality improvement over \pind and \spll is up to~100\%.

\pind and \spll in turn do not perform well, most likely because they use unstable covariance matrices for \pca transformation (note that $\dima > \size$ in all cases tested). At $\dima = 8000$, we have to stop the execution of \pind and \spll due to excessive runtime (more than 12 hours).

\rsif does not perform so well (especially on non-linear correlations), most likely as it assumes that data is uniformly distributed in high dimensional spaces.

\ourmethod in contrast reliably detects changes in high dimensional time series with small window sizes. By not making assumptions on data distributions nor that dimensions are linearly correlated, it copes well with different types of change and yields high quality results.

\subsection{Efficiency Results using Synthetic Data}

\begin{figure}[t]
\centering
\subfigure[Runtime vs.\ dimensionality]
{{\includegraphics[width=0.22\textwidth]{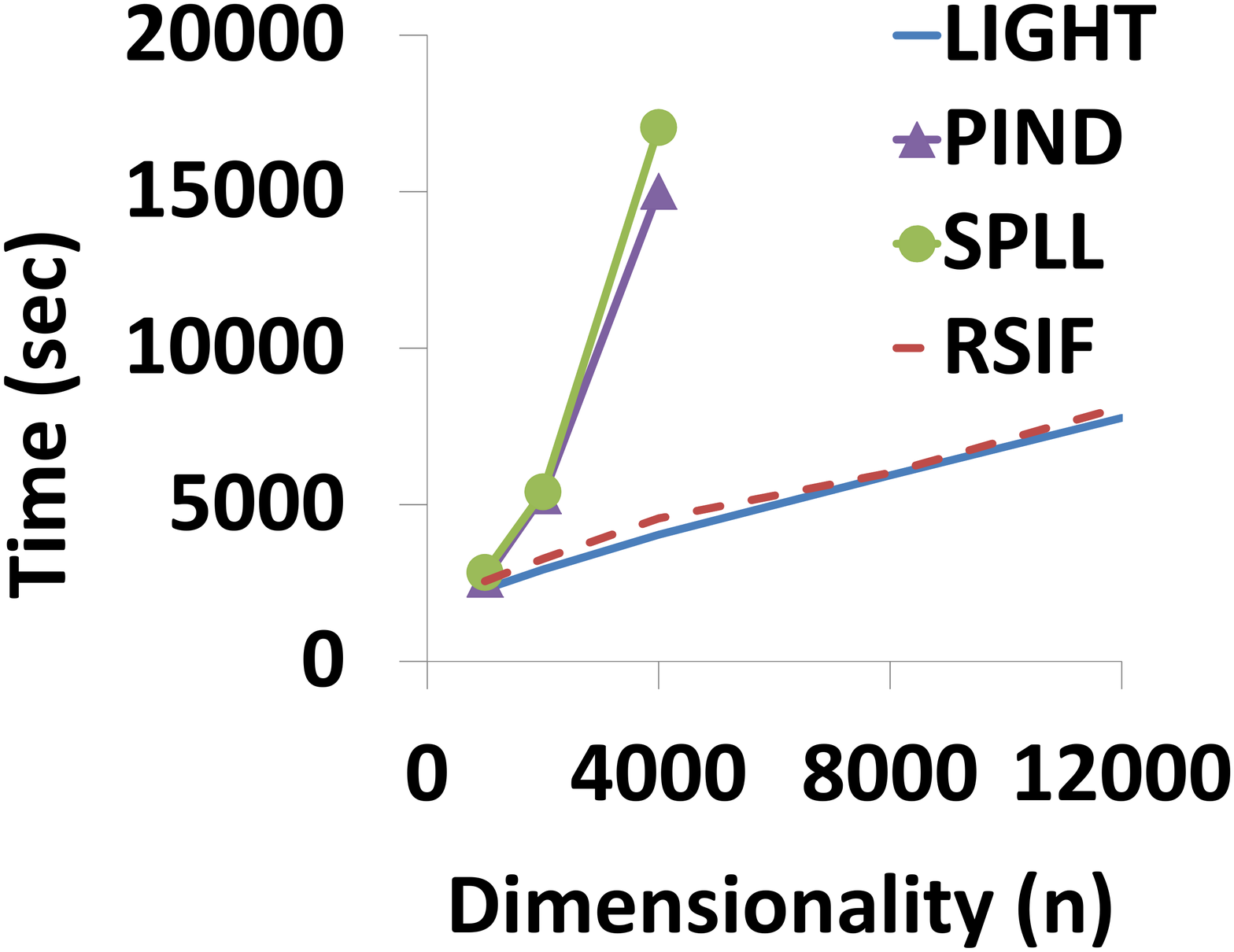}}\label{fig:time_vs_dim_complex}}
\subfigure[Runtime vs.\ dimensionality]
{{\includegraphics[width=0.22\textwidth]{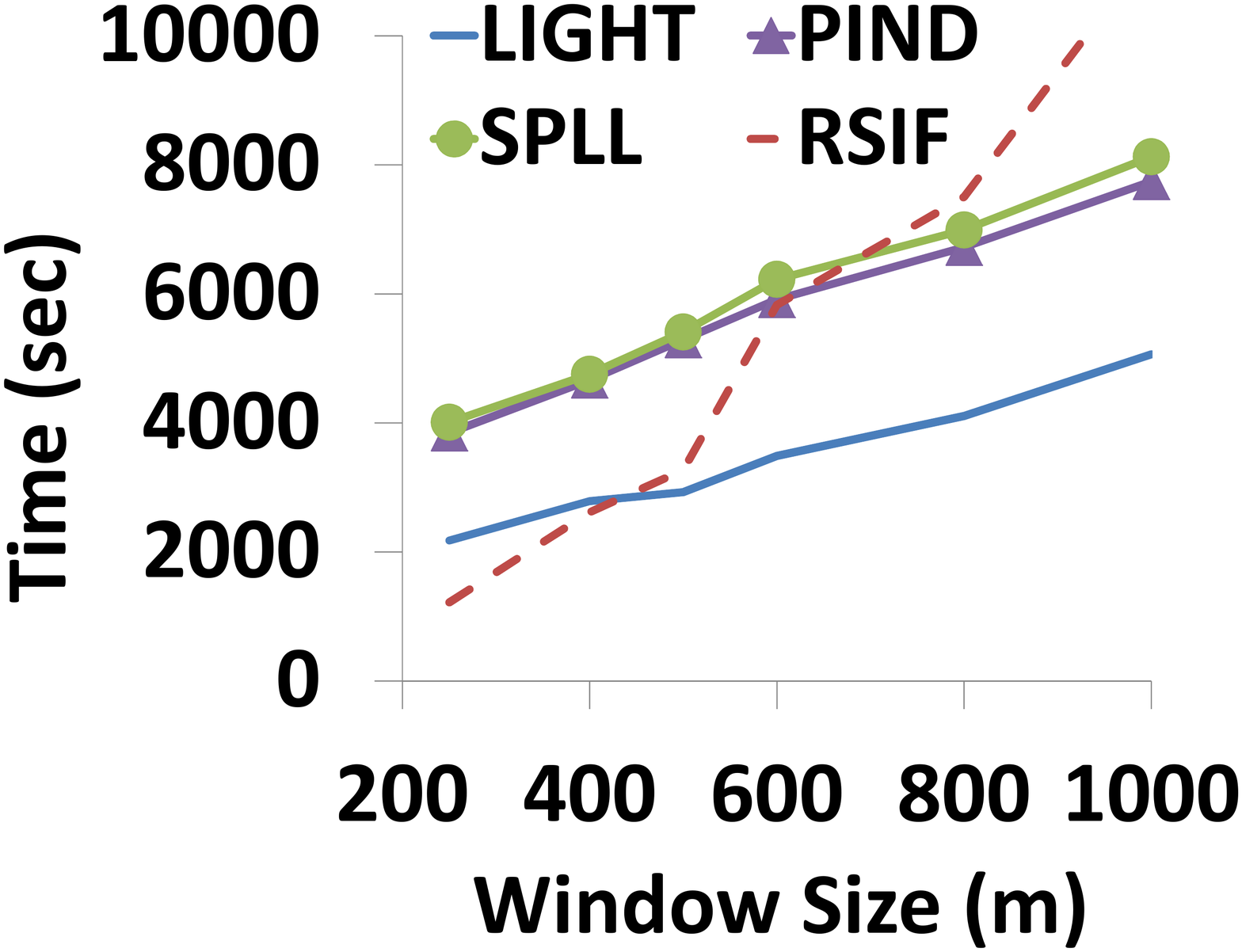}}\label{fig:time_vs_win_complex}}
\caption{[Lower is better] Comparison with competitors: Runtime on synthetic data sets with non-linear correlations. Overall, \ourmethod has the best scalability across different values of dimensionality $\dima$ and window sizes $\size$.}
\end{figure}

Here we test efficiency against window size $\size$ and dimensionality $\dima$. The setup is as above. We show representative results on data sets with \textit{non-linear correlations} in Figures~\ref{fig:time_vs_dim_complex} and~\ref{fig:time_vs_win_complex}. Results on other types of data sets are similar and hence skipped for brevity.

We see that in both cases, \ourmethod is more efficient than all competitors. The performance improvement over \pind and \spll is more than 100\% for high values of $\dima$. Further, the experiments show that \ourmethod has linear scalability to $\dima$ and $\size$, which corroborates our analysis in Section~\ref{sec:summing}. 

\subsection{Parameter Sensitivity}

To assess the sensitivity of \ourmethod to its parameters, we re-use the synthetic data sets above, fixing $\dima = 2000$ and $\size = 500$. The default setting for the parameters is: $c = 200$, percentage $= 90\%$, $s_1 = 50$, and $s_2 = 3$. That is, when testing against one parameter we use the default setting for the other parameters.

\begin{figure}[t]
\centering
\subfigure[F1 and Runtime vs.\ $c$]
{{\includegraphics[width=0.23\textwidth]{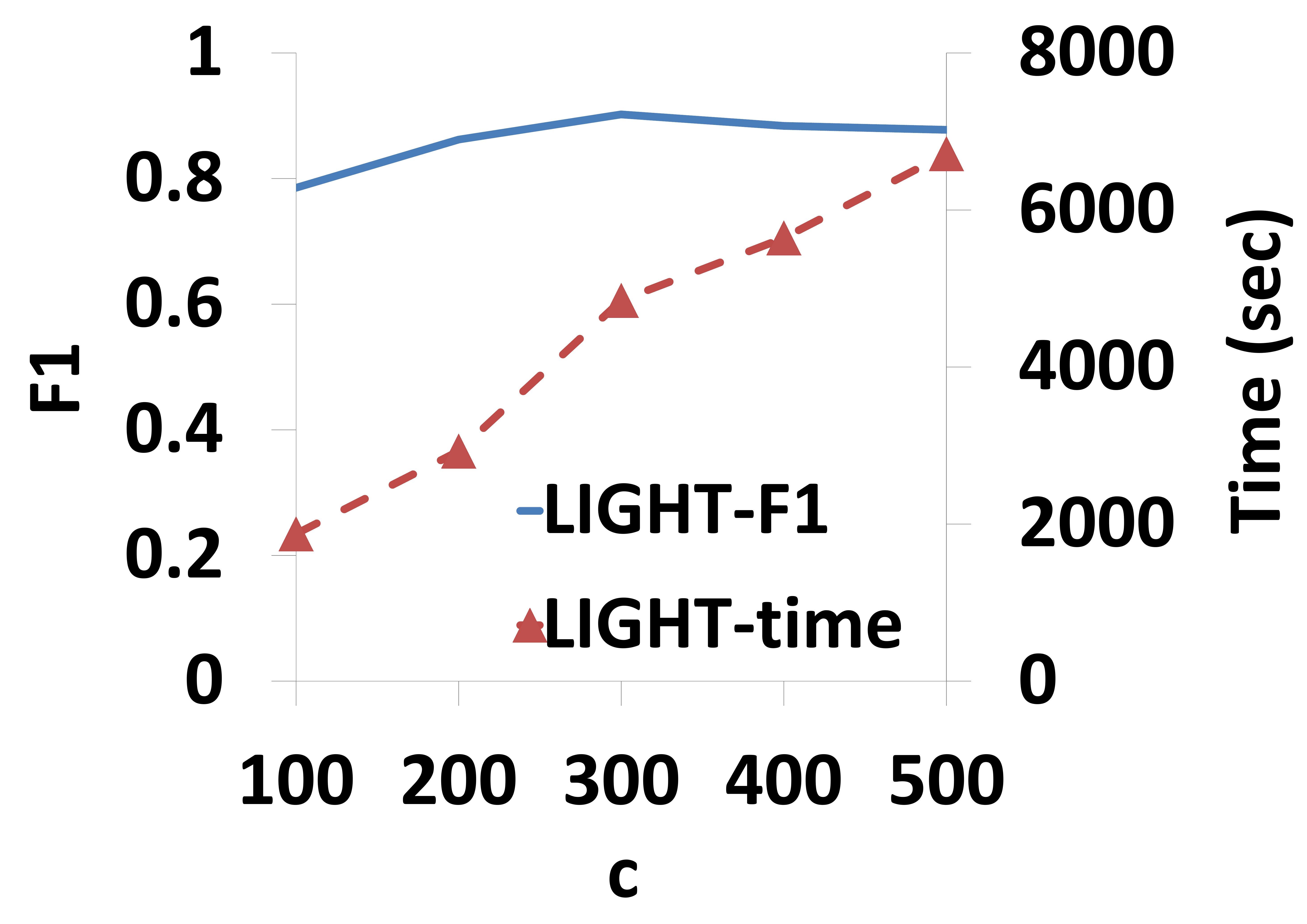}}\label{fig:c_complex}}~
\subfigure[F1 and Runtime vs.\ \% variance]
{{\includegraphics[width=0.23\textwidth]{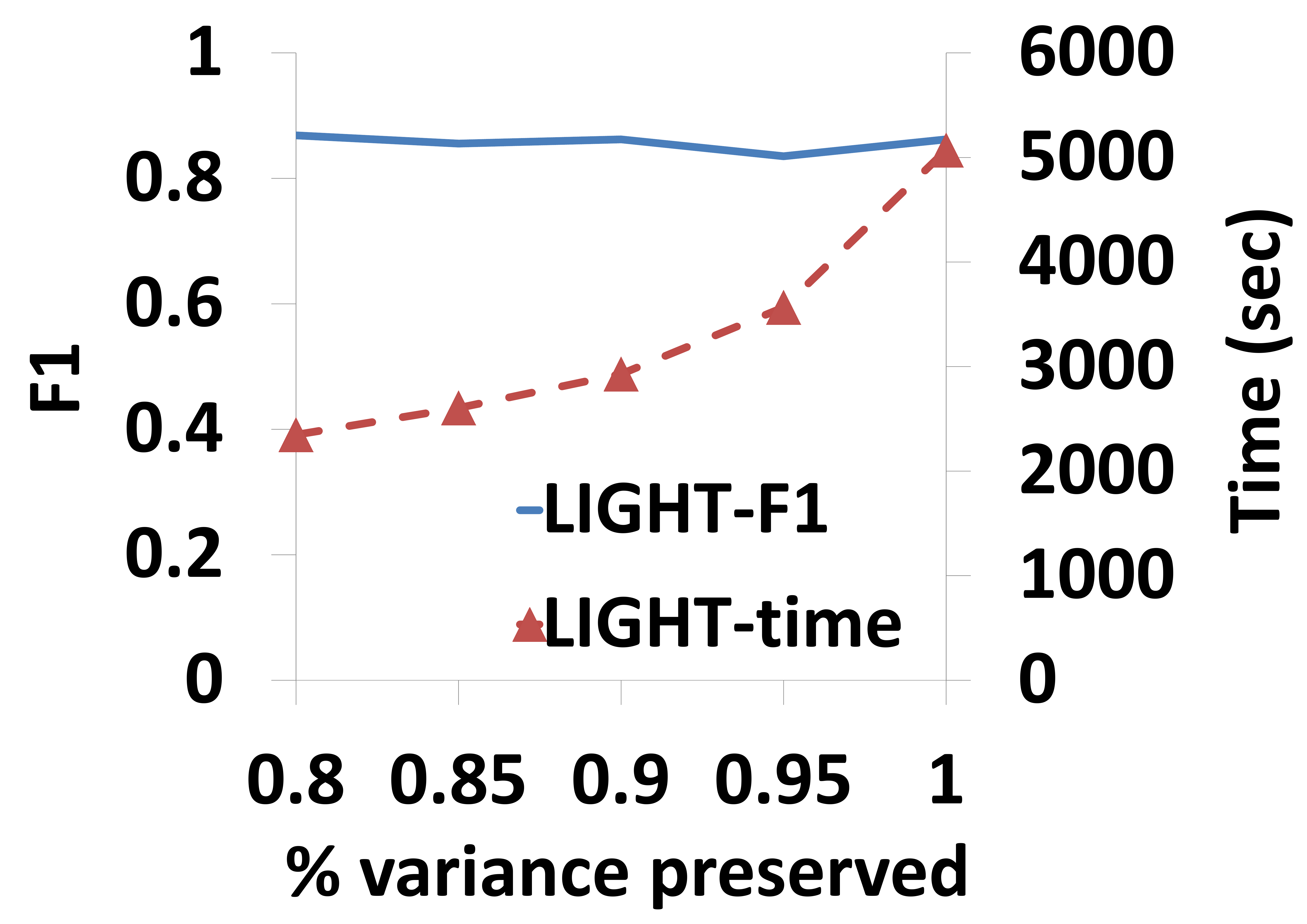}}\label{fig:p_complex}}
\subfigure[F1 and Runtime vs.\ \% $s_1$]
{{\includegraphics[width=0.23\textwidth]{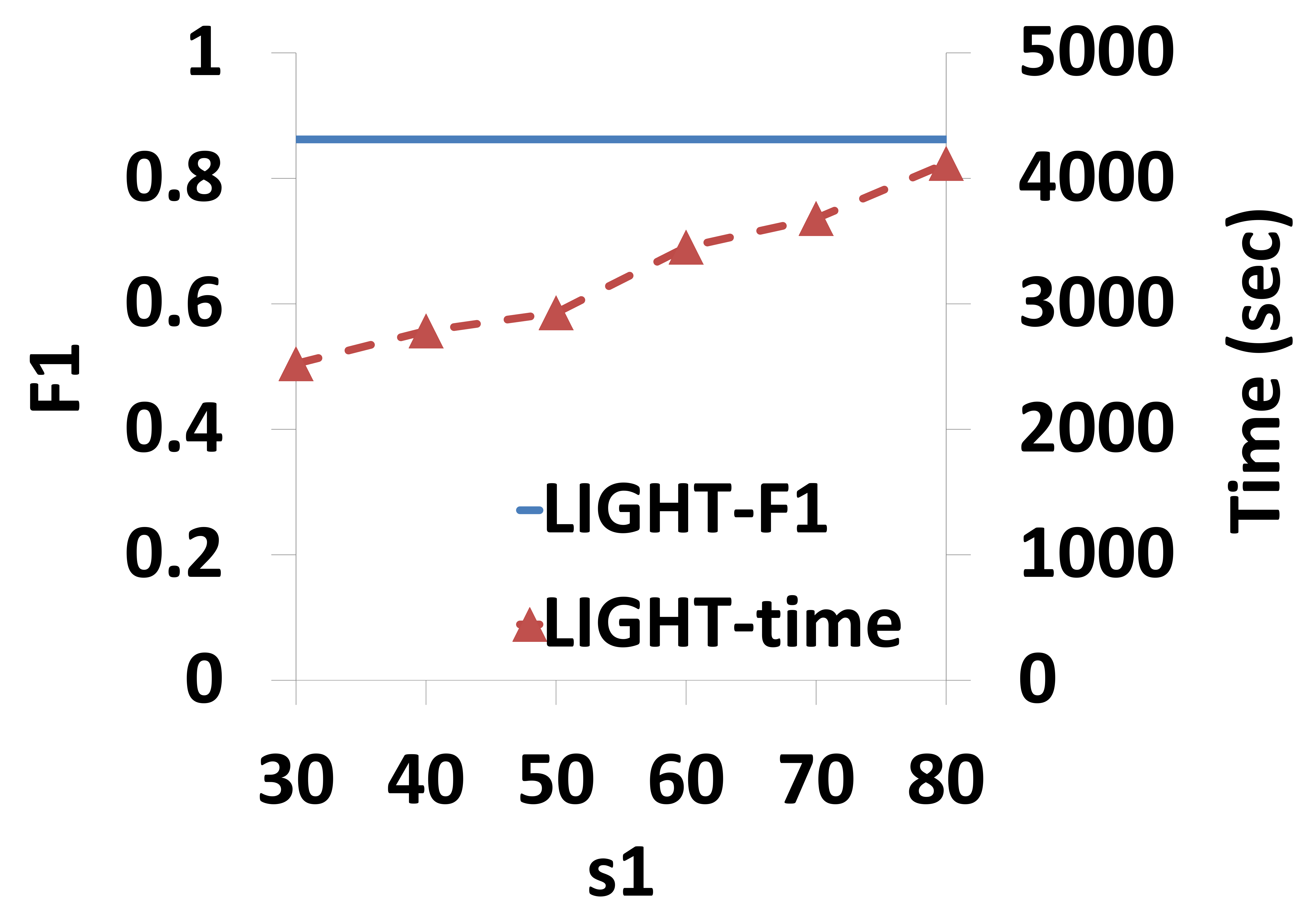}}\label{fig:s1_complex}}~
\subfigure[F1 and Runtime vs.\ \% $s_2$]
{{\includegraphics[width=0.23\textwidth]{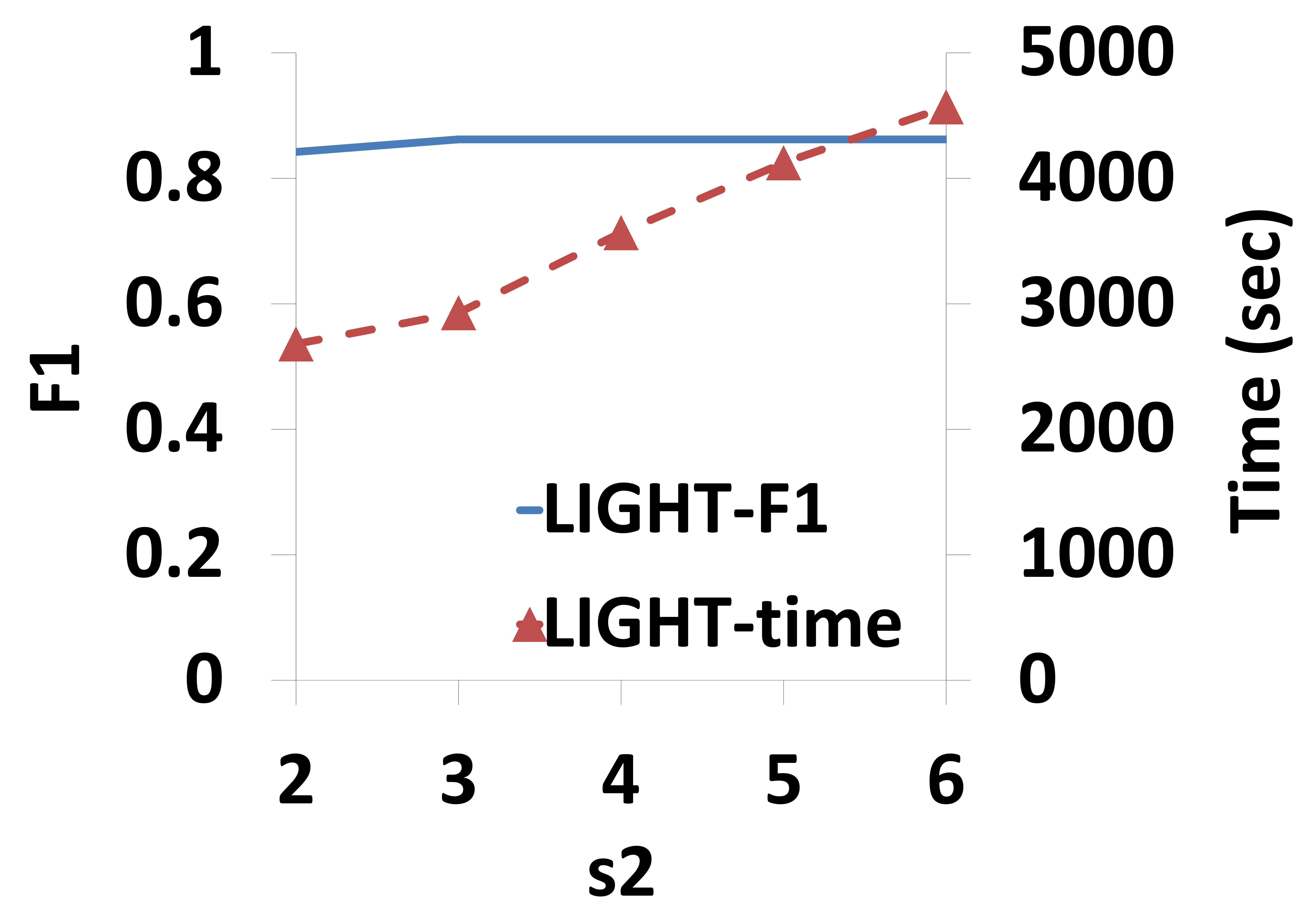}}\label{fig:s2_complex}}
\caption{Sensitivity of \ourmethod to parameter setting on synthetic data sets with non-linear correlations. Overall, in terms of quality \ourmethod is very stable to parameter setting.}
\end{figure}

The representative results on data sets with \textit{non-linear correlations} are in Figures~\ref{fig:c_complex}---\ref{fig:s2_complex}. We see that \ourmethod is very stable to different values of its parameters, which facilitates easy parameterization. The results also suggest that our default setting makes a reasonable choice in terms of both quality and efficiency.

\subsection{Ablation Study}

To study the impact of our design choices, we consider three variants of \ourmethod, each of which we create by switching off one of its properties. The first one is \lind, for \ourmethod with independence assumption. Essentially, \lind applies our scalable \pca mapping (see Section~\ref{sec:ins:transform}) but assumes that dimensions in \pca spaces are statistically independent. It could be seen as an extension of \pind. The second variant is \lnf, for \ourmethod without factorization. That is, \lnf applies our scalable \pca mapping and then computing change score using joint distributions in full \pca spaces. For \lnf, we use the quadratic measure of divergence~\cite{nguyen:ipd} with computation on empirical data in closed form. The third variant is \lnp, for \ourmethod without \pca mapping, i.e.\ factorization is performed on original $\dima$-dimensional spaces.

\begin{figure}[t]
\centering
\subfigure[F1 vs.\ dimensionality]
{{\includegraphics[width=0.22\textwidth]{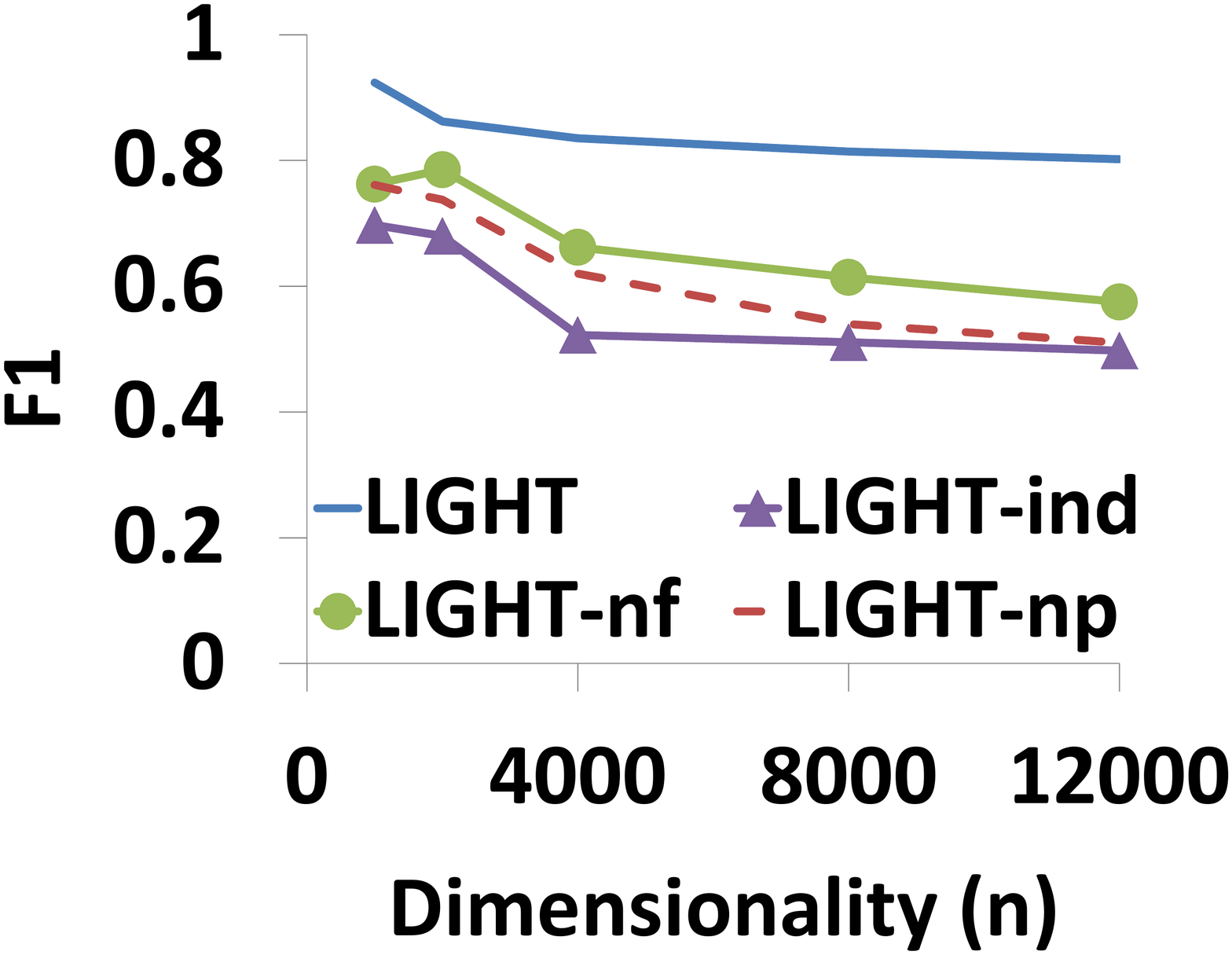}}\label{fig:acc_vs_dim_variants}}
\subfigure[F1 vs.\ dimensionality]
{{\includegraphics[width=0.22\textwidth]{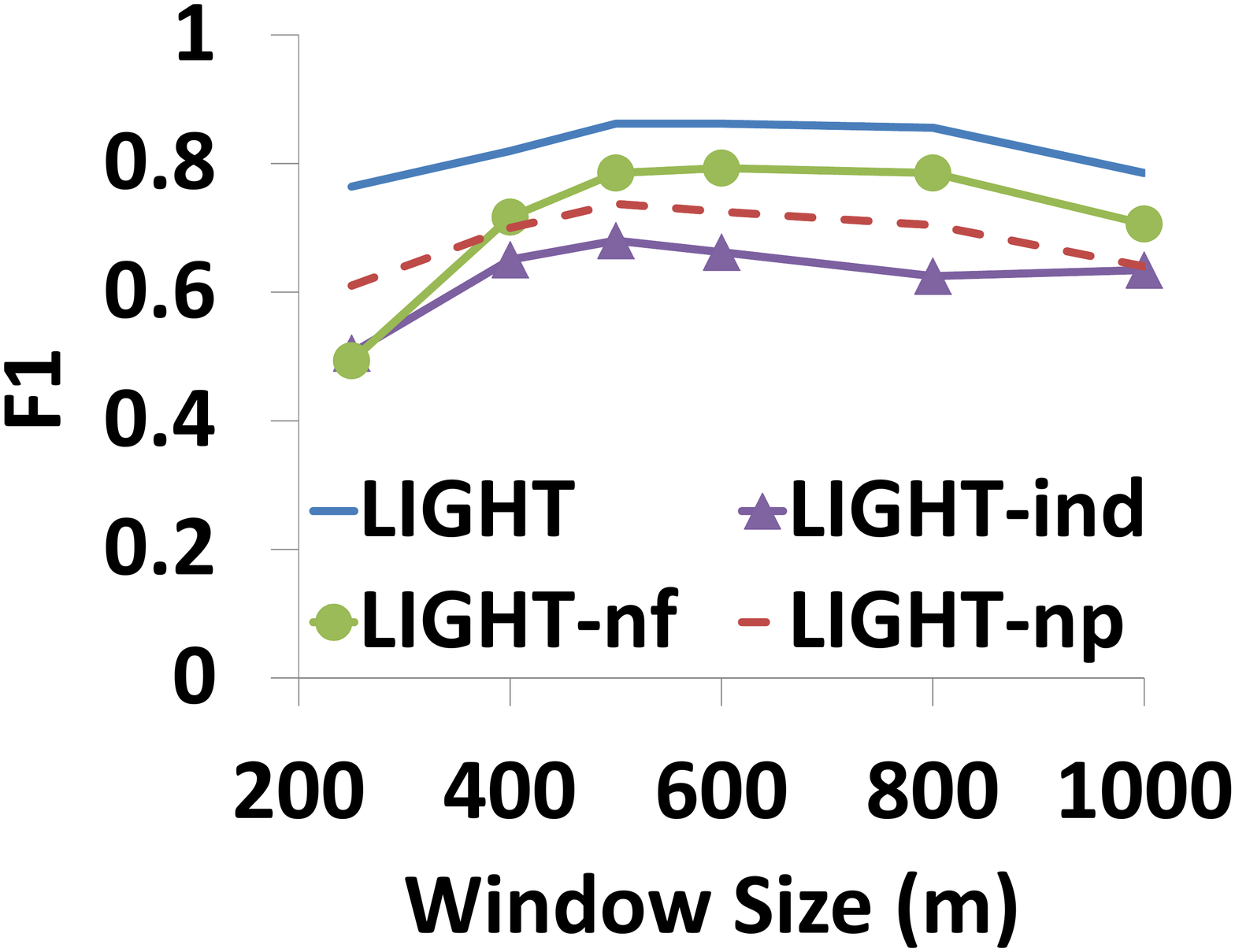}}\label{fig:acc_vs_win_variants}}
\caption{[Higher is better] Comparison with variants: F1 scores on synthetic data sets with non-linear correlations. Overall, \ourmethod outperforms all of its variants.}
\end{figure}

We show representative results on data sets with \textit{non-linear correlations} in Figures~\ref{fig:acc_vs_dim_variants} and~\ref{fig:acc_vs_win_variants}. We can see that \ourmethod outperforms all of its variants. That \ourmethod is better than \lind highlights the importance of taking into account correlations in \pca spaces. That \ourmethod outperforms \lnf shows the effectiveness of our factorization step. Finally, \lnp tries to approximate very high dimensional distributions. This is much harder than approximating lower dimensional ones, as \ourmethod does. This explains why \ourmethod achieves a better performance than \lnp. In terms of runtime, \lind and \lnf are faster than \ourmethod since they do not spend time to factorize the joint distribution of each \pca space. The difference however is negligible.



\subsection{Results on Real Data}

We now study the performance of \ourmethod on real data. We give an overview of the data sets in Table~\ref{tab:datasets}. Each of them is a labeled time series. As change points, we use these class labels, a common practice in change detection literature~\cite{kawa:kliep,song:mixedg}. For Human Activities (HAR) and Human Postural Transitions (HAPT), as their numbers of dimensions are only in the hundreds, we set $c = 50$ and $\size = 100$ for \ourmethod. For the other time series, we use its default parameter setting. We evaluate quality using the F1 measure.

The results are in Table~\ref{tab:real_data_results} and~\ref{tab:real_data_time}. Overall, we can see that \ourmethod consistently yields the best quality and is the most efficient across all time series. Notice that EMG1, EMG2, and Sport are relatively small in length while having many dimensions. For an effective change detection on such time series, it is necessary to use small window sizes. This is an issue for \pind and \spll. In particular, they have to perform \pca transformation on very unstable covariance matrices, which could be an explanation why they do not perform well on EMG1, EMG2, and Sport. For Amazon and Youtube data, the runtime of \pind and \spll exceeds 12 hours. \ourmethod in contrast achieves high accuracy on all high dimensional time series tested. Further, it finishes within 1.5 hours even on the 20\,000 dimensional Amazon and 50\,000 dimensional Youtube data, and has very high accuracy on both data sets.

\begin{table}[t]
\centering 
\begin{tabular}{lrrrr}
\toprule
{\bf Data} & {\bf \ourmethod} & {\bf \pind} & {\bf \spll} & {\bf \rsif}\\
\otoprule

Amazon & \textbf{0.91} & - & - & 0.64\\

EMG1 & \textbf{0.77} & 0.48 & 0.45 & 0.72\\

EMG2 & \textbf{0.84} & 0.41 & 0.44 & 0.67\\

HAR & \textbf{0.83} & 0.62 & 0.55 & 0.70\\

HAPT & \textbf{0.85} & 0.68 & 0.62 & 0.71\\

Sport & \textbf{0.94} & 0.51 & 0.46 & 0.84\\





Youtube & \textbf{0.93} & - & - & 0.76\\

\midrule
		
Average & \textbf{0.87} & 0.54 & 0.50 & 0.72\\

\bottomrule
\end{tabular}
\caption{[Higher is better] F1 scores on real data sets. Best values are in \textbf{bold}. `-' means excessive runtime (more than 12 hours). Overall, \ourmethod consistently yields the best quality across all data sets.} \label{tab:real_data_results} 
\end{table}

\begin{table}[t]
\centering 
\begin{tabular}{lrrrr}
\toprule
{\bf Data} & {\bf \ourmethod} & {\bf \pind} & {\bf \spll} & {\bf \rsif}\\
\otoprule

Amazon & \textbf{1273.6} & $\infty$ & $\infty$ & 1944.5\\

EMG1 & \textbf{1.2} & 92.6 & 98.1 & 3.1\\

EMG2 & \textbf{1.1} & 345.7 & 341.5 & 2.3\\

HAR & \textbf{2.9} & 11.2 & 12.8 & 3.3\\

HAPT & \textbf{2.4} & 12.5 & 12.4 & 3.3\\

Sport & \textbf{5.6} & 1295.7 & 1280.4 & 11.9\\





Youtube & \textbf{4863.5} & $\infty$ & $\infty$ & 7338.4\\

\midrule

Average & \textbf{878.6} & $\infty$ & $\infty$ & 1329.5\\

\bottomrule
\end{tabular}
\caption{[Lower is better] Runtime (in seconds) on real data sets. Best values are in \textbf{bold}. `$\infty$' means excessive runtime (more than 12 hours). Overall, \ourmethod consistently is the most efficient across all data sets.} \label{tab:real_data_time} 
\end{table}

\section{Discussion} \label{sec:dis}

The experiments show that \ourmethod is both very efficient and yields high quality for change detection in very high dimensional time series containing different types of change, be them linear or non-linear. Furthermore, it allows small window sizes to be used. This makes it applicable to different types of time series, e.g.\ those where the dimensionality is even larger than the series length, such as EMG1. Its good performance over the competition suggests the following. First, our scalable \pca transformation is much more effective than traditional \pca mapping when it comes to high dimensionality. The benefits scalable \pca mapping brings here lie in both quality and efficiency. Second, our distribution factorization yields better quality than using joint distributions or assuming statistical independence in \pca spaces. That \ourmethod outperforms competitors and its variants could also be attributed to our new divergence measure, which can capture changes in both linear and non-linear structures.

Yet, there is room for alternative methods as well as further improvements. For instance, in this paper we use \pca for dimension reduction. Other methods, e.g.\ canonical correlation analysis~\cite{chang:cca}, can also be used. Besides the matrix sampling method we employ, it also is interesting to explore other related techniques, such as~\cite{sarlos:sampling,halko:sampling}. The details, however, are beyond the scope of this work. In addition, we here pursue the non-parametric setting. As long as the knowledge on data distributions is known, one can resort to parametric methods to compute other divergence measures, e.g.\ Kullback-Leibler divergence or Jensen-Shannon divergence. As future work, we plan to extend \ourmethod to time series with mixed data types, e.g.\ those with numeric and categorical dimensions. This will help us to enrich the capability of \ourmethod in real-world applications.

\section{Conclusion} \label{sec:con}

In this paper, we studied the problem of change detection on very high dimensional time series. This setting poses both efficiency and quality issues. To address these, we proposed \ourmethod. In short, it works in three steps: 1) scalable \pca mapping to reduce dimensionality, 2) scalable factorization of joint distributions in \pca spaces to increase robustness, and 3) scalable computation of divergence scores on factorized distributions. Experiments on both synthetic and real-world data show \ourmethod outperforms state of the art with up to 100\% improvement in both quality and efficiency.

\section*{Acknowledgements}
The authors are supported by the Cluster of Excellence ``Multimodal Computing and Interaction'' within the Excellence Initiative of the German Federal Government.

\bibliographystyle{abbrv}
\bibliography{bib/abbrev,bib/citation,bib/bib-jilles}

\ifapx
\appendix
\section{Proofs} \label{sec:proofs}

\begin{proof}[Lemma~\ref{lem:klfactor}]
By definition we have that
\begin{align*}
&\kl\left(p(Y_1, \ldots, Y_k)\; ||\; q(Y_1, \ldots, Y_k)\right)\\
&= \int p(y_1, \ldots, y_k) \log \frac{p(y_1, \ldots, y_k)}{q(y_1, \ldots, y_k)} dy_1 \cdots dy_k.
\end{align*}
From Section~\ref{sec:ins:factor} and based on the convention of our framework, we have $p(Y_1, \ldots, Y_k) = \frac{\prod_{(Y_i, Y_j) \in \edge} p(Y_i, Y_j)}{\prod_{Y \in \node} p(Y)^{\degr(Y) - 1}}$ and $q(Y_1, \ldots, Y_k) = \frac{\prod_{(Y_i, Y_j) \in \edge} q(Y_i, Y_j)}{\prod_{Y \in \node} q(Y)^{\degr(Y) - 1}}$. Using these information we deduct that
\begin{align*}
&\kl\left(p(Y_1, \ldots, Y_k)\; ||\; q(Y_1, \ldots, Y_k)\right)\\
&= \sum_{(Y_i, Y_j) \in \edge} \int p(y_1, \ldots, y_k) \log \frac{p(y_i, y_j)}{q(y_i, y_j)} dy_1 \cdots dy_k\\
&- \sum_{Y \in \node: \degr(Y) > 1} \int p(y_1, \ldots, y_k) \log \frac{p(y)}{q(y)} dy.
\end{align*}
Thus, we arrive at the result.
\end{proof}

\begin{proof}[Lemma~\ref{lem:score}]
First, we prove that
\begin{align*}
&\diff\left(P(Y_i)\; ||\; Q(Y_i)\right) \\
&\leq \left(\max(Y_j) - \min(Y_j)\right) \times \diff\left(P(Y_i, Y_j)\; ||\; Q(Y_i, Y_j)\right).
\end{align*}
In particular we have that $P(y_i) = \int P(y_i, y_j) dy_j$ and similarly for $Q(y_i)$. Thus,
\begin{align*}
&\left(P(y_i) - Q(y_i)\right)^2 \\
&= \left(\int \left(P(y_i, y_j) - Q(y_i, y_j)\right) dy_j\right)^2 \\
&\leq \left(\max(Y_j) - \min(Y_j)\right) \times \int \left(P(y_i, y_j) - Q(y_i, y_j)\right)^2 dy_j.
\end{align*}
The inequality is in fact the Cauchy-Schwarz inequality. Hence,
\begin{align*}
&\frac{1}{\left(\max(Y_j) - \min(Y_j)\right)} \times \diff\left(P(Y_i)\; ||\; Q(Y_i)\right) \\
&\leq \int \left(P(y_i, y_j) - Q(y_i, y_j)\right)^2 dy_i dy_j.
\end{align*}
The right hand side of the above inequality is indeed $\diff\left(P(Y_i, Y_j)\; ||\; Q(Y_i, Y_j)\right)$.

As $Y_i$s are obtained by \pca it holds that
$$Y_i \in \left[-\sqrt{\sum\limits_{i=1}^{\dima} \max\{\minv_i^2, \maxv_i^2\}}, \sqrt{\sum\limits_{i=1}^{\dima} \max\{\minv_i^2, \maxv_i^2\}}\right].$$
We now need to prove that for all $Y \in \Yb = \{Y: \degr(Y) > 1\}$, we can choose for each $Y$ a set of $(\degr(Y) - 1)$ terms of the form $\diff\left(P(Y, Y_i)\; ||\; Q(Y, Y_i)\right)$ such that $(Y, Y_i) \in \edge$ and different $Y$s will not share any common term. First, as the number of edges of the maximum spanning tree is $(k - 1)$, we have that
$$\sum_{Y \in \Yb} (\degr(Y) - 1) = k - 2.$$
Now we pick the terms for all $Y \in \Yb$ as follows. We consider edges $(Y_i, Y_j)$ whose one end-point is a leaf. When $k > 2$ it must be the case that either $Y_i \in \Yb$ or $Y_j \in \Yb$. Assuming that the former holds we remove edge $(Y_i, Y_j)$ and assign term $\diff\left(P(Y_i, Y_j)\; ||\; Q(Y_i, Y_j)\right)$ to $Y_i$. We carry on until all edges are removed. Note that under this procedure a node $Y \in \Yb$ is not removed from the tree until $\degr(Y)$ becomes~1. This also means that when $Y$ is removed there have been $(\degr(Y) - 1)$ terms assigned to it. This completes the second part of the proof.

With the results in the first part and the second part, we arrive at $\score \geq 0$. Equality happens when $P(Y_i, Y_j) = Q(Y_i, Y_j)$ for $(Y_i, Y_j) \in \edge$ and $P(Y) = Q(Y)$ for $Y \in \Yb$. This means that $p(Y_1, \ldots, Y_k)$ and $q(Y_1, \ldots, Y_k)$ under the factorization model are equal.

When $p(Y_1, \ldots, Y_k)$ and $q(Y_1, \ldots, Y_k)$ are equal, we have that $P(Y_i, Y_j) = Q(Y_i, Y_j)$ for $(Y_i, Y_j) \in \edge$ and $P(Y) = Q(Y)$ for $Y \in \Yb$. Thus, $\score = 0$. We complete our proof. 
\end{proof}

\section{Incremental Computation of Divergence Score} \label{sec:incremental}

We illustrate the idea on incrementally computing $\diff\left(p(E, F)\; ||\; q(E, F)\right)$ where $E, F \in \{Y_1, \ldots, Y_k\}$. Incrementally computing $\diff\left(p(Y)\; ||\; q(Y)\right)$ where $Y \in \{Y_1, \ldots, Y_k\}$ follows straightforwardly.

Assume that the empirical data forming $p(E, F)$ contains data points $\{(e_{p,1}, f_{p,1}), \ldots, (e_{p,\size}, f_{p,\size})\}$. Analogously, we denote $\{(e_{q,1}, f_{q,1}), \ldots, (e_{q,\size}, f_{q,\size})\}$ as the data points forming $q(E, F)$. According to~\cite{nguyen:ipd}, $\diff\left(p(E, F)\; ||\; q(E, F)\right) =$
\begin{align*}
&\frac{1}{\size^2} \sum_{j_1=1}^{\size}\sum_{j_2=1}^{\size} \left(\maxv_e - \max(e_{p,j_1}, e_{p,j_2})\right) \left(\maxv_f - \max(f_{p,j_1}, f_{p,j_2})\right) \\
&- \frac{2}{\size^2} \sum_{j_1=1}^{\size}\sum_{j_2=1}^{\size}\left(\maxv_e - \max(e_{p,j_1}, e_{q,j_2})\right) \left(\maxv_f - \max(f_{p,j_1}, f_{q,j_2})\right) \\ 
&+ \frac{1}{\size^2} \sum_{j_1=1}^{\size}\sum_{j_2=1}^{\size} \left(\maxv_e - \max(e_{q,j_1}, e_{q,j_2})\right) \left(\maxv_f - \max(f_{q,j_1}, f_{q,j_2})\right).
\end{align*}
Thus, $\diff\left(p(E, F)\; ||\; q(E, F)\right)$ can be factorized per data point. By storing the contribution of each point to $\diff\left(p(E, F)\; ||\; q(E, F)\right)$, we can incrementally update this score in $O(\size)$ time. We have $O(k)$ score terms in total. Thus, the cost to compute divergence score for each new sample of the time series is $O(\size k)$.

\section{Scaling to Large Window Sizes} \label{sec:scalinglarge}

The idea is that $\score$ is computed based on terms of the form $\diff\left(p(.)\; ||\; q(.)\right)$ where $p(.)$ and $q(.)$ are estimated from the data of $\wreft$ and $\wtestt$, respectively. An implicit assumption here is that the data of $\wreft$ (similarly for $\wtestt$) are i.i.d.\ samples of $p(.)$. By definition, i.i.d.\ samples are obtained by randomly sampling from an infinite population or by randomly sampling with replacement from a finite population. In both cases, the distribution of i.i.d.\ samples are assumed to be identical to the distribution of the population. This is especially true when the sample size is very large~\cite{scott:density}. Thus, when $\size$ is very large the empirical distribution $\hat{p}(.)$ formed by $\wreft$ approaches the true distribution $p(.)$. Assume now that we randomly draw with replacement $\epsilon \times \size$ samples of $\wreft$ where $\epsilon \in (0, 1)$. As mentioned above, these subsamples contain i.i.d.\ samples of $\hat{p}(.) \approx p(.)$. As with any set of i.i.d.\ samples with a reasonable size, we can assume that the empirical distribution formed by the subsamples is identical to $p(.)$. Thus, we can use them to approximate $\score$. In that case, the complexity of computing $\score$ for initial $\wref$ and $\wtest$ or after every change is reduced to $O(\epsilon^2 \size^2 k)$. When subsampling is needed we only use it once for each $\wref$.

\fi

\end{document}